\documentclass[11pt, english]{article}
\usepackage{babel}
\usepackage[latin1]{inputenc}
\usepackage[T1]{fontenc}
\usepackage{amsmath}
\usepackage{graphicx}
\usepackage{amssymb}
\usepackage{amsthm}
\usepackage{subcaption}
\usepackage{bbm}
\usepackage{mathtools}

\usepackage{xr}

\usepackage{algorithmic}
\usepackage{courier} 
\usepackage{algorithm}
\graphicspath{{figures/}}

\usepackage{chngcntr}
\usepackage{apptools}
\AtAppendix{\counterwithin{thm}{section}}

\def\^{\widehat}

\newcommand{\norm}[1]{\Vert #1 \Vert}

\def\phi{\varphi}

\numberwithin{equation}{section}

\renewcommand{\phi}{\varphi}

\def\~{\widetilde}
\def\^{\widehat}
\newcommand{\ee}{{\rm e}\hspace{1pt}}

\newcommand{\dd}{\hspace{1pt}{\rm d}\hspace{0.5pt}}
\newcommand{\abs}[1]{\left| #1 \right|}

\newcommand{\veps}{\varepsilon}

\newtheorem{thm}{Theorem}

\newtheorem{lem}[thm]{Lemma}
\newtheorem{cor}[thm]{Corollary}
\newtheorem{defn}[thm]{Definition}

\title{Computing Tight Differential Privacy Guarantees Using FFT}

\author{Antti Koskela$^{1}$, Joonas J\"alk\"o$^{2}$ and Antti Honkela$^3$ \vspace{5mm} \\ 
$^1$ Helsinki Institute for Information Technology HIIT,\\
    Department of Mathematics and Statistics, University of Helsinki, Finland \\
  $^2$ Helsinki Institute for Information Technology HIIT,\\
    Department of Computer Science, Aalto University, Finland \\
  $^3$ Helsinki Institute for Information Technology HIIT,\\
    Department of Computer Science, University of Helsinki}

    \date{}

\begin{document}
	
\maketitle

\abstract{
   Differentially private (DP) machine learning has recently become
   popular. 
   The privacy loss of DP algorithms is commonly reported
   using $(\varepsilon,\delta)$-DP. In this paper, we propose
   a numerical accountant for evaluating the privacy loss for
   algorithms with continuous one dimensional output. This accountant
   can be applied to the subsampled multidimensional Gaussian mechanism 
   which underlies the popular DP stochastic gradient descent.
   The proposed method is based on a numerical approximation 
   of an integral formula which gives the exact $(\varepsilon,\delta)$-values. 
   The approximation is carried out by discretising the integral and by evaluating discrete convolutions using the fast Fourier transform algorithm. 
   We give both theoretical error bounds and numerical error estimates for the approximation.
   Experimental comparisons with state-of-the-art techniques demonstrate
   significant improvements in bound tightness and/or computation time.
   Python code for the method can be found in Github.\footnote{https://github.com/DPBayes/PLD-Accountant/}
}

\section{Introduction}

Differential privacy (DP) \cite{dwork_et_al_2006} has clearly been
established as the dominant paradigm for privacy-preserving machine
learning. Early work on DP machine learning focused on single shot
perturbations for convex problems (e.g.~\cite{Chaudhuri2011}), while
contemporary research has focused on iterative algorithms such as DP
stochastic gradient descent (SGD) \cite{Rajkumar2012,Song2013,Abadi2016} .

Evaluating the privacy loss of an iterative algorithm is based on the
composition theory of DP. The so-called advanced composition theorem
of \cite{dwork2010} showed how to trade decreased
$\varepsilon$ with slightly increased $\delta$ in
$(\epsilon, \delta)$-DP. This was further improved
e.g. by \cite{kairouz2017}. The privacy amplification
by subsampling \cite{chaudhuri06, Beimel2013, Bassily2014, wang15privacy}
is another component that has been studied to improve the privacy
bounds.

A major breakthrough in obtaining tighter composition bounds came from
using the entire privacy loss profile of DP algorithms instead of
single $(\varepsilon, \delta)$-values. This was first introduced by the moments
accountant~\cite{Abadi2016}. 
The development of R\'enyi differential privacy
(RDP) \cite{mironov2017} allowed tight bounds on the privacy cost of
composition, and recently proposed amplification theorems for RDP~\cite{balle2018subsampling,wang2019} showed how subsampling affects the privacy cost of
RDP. In~\cite{zhu2019} tight RDP bounds were given for the Poisson subsampling method.

Using the recently introduced privacy loss distribution (PLD) formalism
\cite{sommer2019privacy}, we compute tight $(\varepsilon, \delta)$-DP bounds on the
composition of subsampled Gaussian mechanisms, using discrete Fourier
transforms to evaluate the required convolutions. We show numerically
that the achieved privacy bounds are tighter than
those obtained by R\'enyi DP compositions and the moments accountant.

Within this computational framework, in addition to the commonly considered Poisson subsampling method, 
we are also able to compute tight privacy bounds for 
the subsampling with replacement and subsampling without replacement methods. 

\section{Differential Privacy}

We first recall some basic definitions of differential privacy~\cite{DworkRoth}. We use the
following notation. An input dataset containing $N$ data points is denoted as $X = (x_1,\ldots,x_N)
\in \mathcal{X}^N$, where $x_i \in \mathcal{X}$, $1 \leq i \leq N $.

\begin{defn} \label{def:adjacency}
	We say two datasets $X$ and $Y$ are neighbours in remove/add relation if you get 
	one by removing/adding an element from/to to other and denote it with $\sim_R$.
	We say $X$ and $Y$ are neighbours in substitute relation if you get one by substituting
	one element in the other. We denote this with $\sim_S$.
\end{defn}

\begin{defn} \label{def:dp}
	Let $\varepsilon > 0$ and $\delta \in [0,1]$. Let $\sim$ define a neighbouring relation.
	Mechanism $\mathcal{M} \, : \, \mathcal{X}^N \rightarrow \mathcal{R}$ is $(\varepsilon,\delta,\sim)$-DP 
	if for every $X \sim Y$
	and every measurable set $E \subset \mathcal{R}$ it holds that
	$$
		\mathrm{Pr}( \mathcal{M}(X) \in E ) \leq \ee^\varepsilon \mathrm{Pr} (\mathcal{M}(Y) \in E ) + \delta.
	$$
	When the relation is clear from context or irrelevant, we will abbreviate it as $(\veps, \delta)$-DP. 
	We call $\mathcal{M}$ tightly $(\veps,\delta,\sim)$-DP, if there does not exist $\delta' < \delta$
	such that $\mathcal{M}$ is $(\veps,\delta',\sim)$-DP.
\end{defn}

\section{Privacy loss distribution} \label{sec:pld}

We first introduce the basic tool for obtaining tight privacy bounds: the privacy loss distribution (PLD).

The results in this section can be seen as continuous versions of their discrete counterparts given in~\cite{meiser2018tight} and~\cite{sommer2019privacy}.
Detailed proofs are given in Appendix. The results apply for both neighbouring relations $\sim_S$ and $\sim_R$. 

We consider mechanisms $\mathcal{M} \, : \, \mathcal{X}^N \rightarrow \mathbb{R}$ which give
as an output distributions with support equaling $\mathbb{R}$. 

\begin{defn} \label{def:plf}
Let $\mathcal{M} \, : \, \mathcal{X}^N \rightarrow \mathbb{R}$ be a randomised mechanism and
let $X \sim Y$. Let $f_X(t)$ denote the density function of
$\mathcal{M}(X)$ and $f_Y(t)$ the density function of $\mathcal{M}(Y)$. Assume
$f_X(t)>0$ and $f_Y(t) > 0$ for all $t \in \mathbb{R}$.
 We define the privacy loss function
of $f_X$ over $f_Y$ as
$$
\mathcal{L}_{X/Y}(t)  = \log \frac{f_X(t)}{f_Y(t)}.
$$
\end{defn}

The following gives the definition of the privacy loss distribution via its density function.
We note that the assumptions on differentiability and bijectivity of the privacy loss function
hold for the subsampled Gaussian mechanism which is considered in Sec.~\ref{Sec:subsampled_gauss}.

\begin{defn} \label{def:pld}
	Let the assumptions of Def.~\ref{def:plf} hold and
	 suppose $\mathcal{L}_{X/Y} \, : \,\mathbb{R} \rightarrow D$, $D \subset \mathbb{R}$ is a continuously differentiable bijective function.
 The privacy loss distribution (PLD) of $\mathcal{M}(X)$ over $\mathcal{M}(Y)$ is defined to be a random variable
which has the density function 
\begin{equation*}
	\omega_{X/Y}(s)= \begin{cases}
		 f_X\big( \mathcal{L}_{X/Y}^{-1}(s)  \big)  \,  \frac{\dd \mathcal{L}_{X/Y}^{-1}(s)}{\dd s}, & s \in \mathcal{L}_{X/Y}(\mathbb{R}),\\
		 0, &\mathrm{else.}
	\end{cases}
\end{equation*}
\end{defn}

For the discrete valued versions of the following result, see \cite[Lemmas 5 and 10]{sommer2019privacy}.
\begin{lem}  \label{lem:maxrepr}
Assume $(\veps,\infty) \subset \mathcal{L}_{X/Y}(\mathbb{R})$.
$\mathcal{M}$  is tightly $(\varepsilon,\delta)$-DP for 
$$
\delta(\veps) = \max \{ \delta_{X/Y}(\veps), \delta_{Y/X}(\veps) \}, 
$$
where
\begin{equation*}
	\begin{aligned}
		\delta_{X/Y}(\veps) = \int_\veps^\infty  (1-\ee^{\veps - s}) \, \omega_{X/Y}(s)   \, \dd s, 
	\end{aligned}
\end{equation*}
and similarly for $\delta_{Y/X}(\veps)$.
\end{lem}

The PLD formalism is essentially based on Lemma~\ref{lem:tight_d} which
states that the mechanism $\mathcal{M}$ is tightly $(\veps,\delta)$-DP with
\begin{equation*} 
	\begin{aligned}
\delta(\veps) = \max_{X \sim Y} \Bigg\{ \int\nolimits_\mathbb{R}  \max \{  f_X(t) - \ee^\veps f_Y(t) ,0  \} \, \dd t,  
 \int\nolimits_\mathbb{R}  \max \{  f_Y(t) - \ee^\veps f_X(t) ,0  \} \, \dd t \Bigg\}.
	\end{aligned}
\end{equation*}
The integral representation of Lemma~\ref{lem:maxrepr} is then obtained by change of variables.
Denoting $s=\mathcal{L}_{X/Y}(t)$, it clearly holds that $f_Y(t) = \ee^{-s} f_X(t)$ and
\begin{equation*}
	\begin{aligned}
&	  \max \{  f_X(t) - \ee^\veps f_Y(t) ,0  \} &  \\ & \quad = \begin{cases}
			(1-\ee^{\veps - s}) f_X(t), &\text{ if } s > \veps,\\
				0, &\text{ otherwise.}
		\end{cases}
	\end{aligned}
\end{equation*}
By change of variables $t = \mathcal{L}^{-1}_{X/Y}(s)$, we obtain the representation of Lemma~\ref{lem:maxrepr}.

We get the tight privacy guarantee for compositions from a continuous counterpart of \cite[Thm.\;1]{sommer2019privacy}.
\begin{thm} \label{thm:integral}
Consider $k$ consecutive applications of a mechanism $\mathcal{M}$. Let $\veps > 0$. The composition is tightly $(\veps,\delta)$-DP for $\delta$ given by
$\delta(\veps) = \max \{ \delta_{X/Y}(\veps), \delta_{Y/X}(\veps) \}$, where
\begin{equation*} 
	\begin{aligned}
		\delta_{X/Y}(\veps) = 
		\int_\veps^\infty (1 - \ee^{\veps - s})\left(\omega_{X/Y} *^k \omega_{X/Y} \right) (s)  \, \dd s,
	\end{aligned}
\end{equation*}
where $\omega_{X/Y} *^k \omega_{X/Y}$ denotes the $k$-fold convolution of $\omega_{X/Y}$ (a similar formula holds for $\delta_{Y/X}(\veps)$).
\end{thm}

\section{The discrete Fourier transform}

The discrete Fourier transform $\mathcal{F}$ and its inverse $\mathcal{F}^{-1}$ are linear operators $\mathbb{C}^n \rightarrow \mathbb{C}^n$
that decompose a complex vector into a Fourier series, or reconstruct it from its Fourier series.
Suppose $x = (x_0,\ldots,x_{n-1}), w = (w_0,\ldots,w_{n-1}) \in \mathbb{R}^n$. Then, $\mathcal{F}$ and $\mathcal{F}^{-1}$ are defined as~\cite{stoer_book}
\begin{equation*} 
	\begin{aligned}
	(\mathcal{F} x)_k & = \sum\nolimits_{j=0}^{n-1} x_j \ee^{- i 2 \pi k j / n},  \\
	(\mathcal{F}^{-1} w  )_k  & = \frac{1}{n} \sum\nolimits_{j=0}^{n-1} w_j \ee^{ i 2 \pi k j / n}.
	\end{aligned}
\end{equation*}
Evaluating $\mathcal{F} x$ and $\mathcal{F}^{-1}w$ takes $O(N^2)$ operations,
however evaluation via the Fast Fourier Transform (FFT) \cite{cooley1965} reduces
the computational cost to $O(N \log N)$.

The convolution theorem \cite{stockham1966} states that for periodic discrete convolutions 
it holds that
\begin{equation} \label{eq:conv_thm}
	\sum\nolimits_{i=0}^{n-1} v_i w_{k-i} = \mathcal{F}^{-1} ( \mathcal{F} v \odot \mathcal{F} w),
\end{equation}
where $\odot$ denotes the elementwise product of vectors and the summation indices are  modulo $n$.

\section{Description of the method}

We next describe the numerical method for computing tight DP-guarantees for continuous one dimensional
distributions. 

\subsection{Truncation of convolutions}

We first approximate the convolutions on a truncated interval $[-L,L]$ as
\begin{equation*}
	\begin{aligned}
	(\omega * \omega )(x) 	
	\approx \int_{-L}^L \omega(t) \omega(x-t) \, \dd t =: (\omega \circledast \omega )(x).
\end{aligned}
\end{equation*}
To obtain periodic convolutions for the discrete Fourier transform we need to periodise $\omega$.
Let $\widetilde{\omega}$ be a $2 L$-periodic extension of $\omega$ such that  
$\widetilde{\omega} (t + n 2L ) = \omega(t)$ for all $t \in [-L,L)$ and $n \in \mathbb{Z}$.
We further approximate
\begin{equation} \label{eq:1}
	\begin{aligned}
		\int_{-L}^L \omega(t) \omega(x-t) \, \dd t  \approx \int_{-L}^{L} \widetilde{\omega}(t) \widetilde{\omega}(x-t) \, \dd t. 
	\end{aligned}
\end{equation}

\subsection{Discretisation of convolutions} \label{subsec:discretisation}

Divide the interval $[-L,L]$ on $n$ equidistant points $x_0,\ldots,x_{n-1}$ such that
$$
x_i = -L + i \Delta x, \textrm{ where } \Delta x = 2L/n. 
$$
Consider the vectors 
$$
\boldsymbol{\omega} = \begin{bmatrix} \omega_0 \\ \vdots \\ \omega_{n-1} \end{bmatrix} \quad \textrm{and} \quad
\boldsymbol{\widetilde{\omega}} = \begin{bmatrix} \widetilde{\omega}_0 \\ \vdots \\ \widetilde{\omega}_{n-1} \end{bmatrix},
$$
where 
$$
\omega_i = \omega(-L + i \Delta x) \quad \textrm{and} \quad \widetilde{\omega}_i = \widetilde{\omega}( i  \Delta x).
$$
Assuming $n$ is even, from the periodicity it follows that 
$$
\boldsymbol{\widetilde{\omega}} = D \boldsymbol{\omega}, \quad \textrm{where} \quad D = \begin{bmatrix} 0 & I_{n/2} \\ I_{n/2} & 0 \end{bmatrix}.
$$
We approximate \eqref{eq:1} using a Riemann sum and the convolution theorem \eqref{eq:conv_thm} as 
\begin{equation*}
	\begin{aligned}
(\widetilde{\omega} \circledast \widetilde{\omega} )(i\Delta x) = & \int_{-L}^{L} \widetilde{\omega}(t) \widetilde{\omega}(i \Delta x - t) \, \dd t \\
 \quad \approx  & \Delta x \sum\nolimits_{\ell = 0}^{n-1} \widetilde{\omega}_\ell \, \widetilde{\omega}_{i - \ell } \textrm{ (indices modulo $n$) }  \\
  =  & \Delta x \left[ \mathcal{F}^{-1} \big(\mathcal{F}( \boldsymbol{\widetilde{\omega}} ) \odot  \mathcal{F} ( \boldsymbol{\widetilde{\omega}} )  \big) \right]_i.
	\end{aligned}
\end{equation*}
Discretisation of $k$-fold truncated convolutions leads to $k$-fold discrete convolutions and to the approximation 
\begin{equation*}
	\begin{aligned}
&  (\widetilde{\omega} \circledast^k \widetilde{\omega} )(-L + i \Delta x) \\
& \quad \approx   (\Delta x)^{k-1} \left[D \, \mathcal{F}^{-1} \big(\mathcal{F}( \boldsymbol{\widetilde{\omega}} )^{\odot k}   \big) \right]_i  \\
& \quad =   (\Delta x)^{-1} \left[D \, \mathcal{F}^{-1} \big(\mathcal{F}( D \boldsymbol{\omega} \Delta x)^{ \odot k}   \big) \right]_i,
	\end{aligned}
\end{equation*}
where $^{\odot k}$ denotes $k$th elementwise power of vectors.

\subsection{ Approximation of the $\delta(\veps)$-integral} \label{subsec:desc}

Finally, using the discretised convolutions we approximate the integral formula for the exact $\delta$-value.
Denote the discrete convolution vector
$$
C^k = (\Delta x)^{-1} \left[D \, \mathcal{F}^{-1} \big(\mathcal{F}( D \boldsymbol{\omega} \Delta x)^{\odot k}   \big) \right]
$$
and the starting point of the discrete sum
$$
\ell_\veps = \min \{ \ell \in \mathbb{Z} \, : \, -L + \ell \Delta x > \veps \}.
$$ 
Using the vector $C^k = \begin{bmatrix} C_0^k & \ldots & C_{n-1}^k \end{bmatrix}^T$, we approximate
the integral formula given in Thm.~\ref{thm:integral} as a Riemann sum:
\begin{equation} \label{eq:approx_int}
	\begin{aligned}
	\delta( \veps) & =  \int_\veps^\infty (1 - \ee^{\veps - s})(\omega *^k \omega ) (s)  \, \dd s  \\
	& \approx  \Delta x \sum\nolimits_{\ell=\ell_\veps}^{n-1}  \big(1 - \ee^{\veps - ( - L + \ell \Delta x)} \big) C_\ell^k.
	\end{aligned}
\end{equation}

We call this method the Fourier Accountant (FA) and describe it in the pseudocode of Algorithm~\ref{alg:delta}. 
We give in Sec.~\ref{sec:err_est} error estimates to
determine the parameters $L$ and $n$ 
such that the error caused by approximations is below a desired level.

\begin{algorithm}[ht!]
\caption{Fourier Accountant algorithm}
\begin{algorithmic}
\STATE{Input: privacy loss distribution $\omega$,
number of compositions $k$, truncation parameter $L$, number of discretisation points $n$.}
\vspace{2mm}
\STATE{Evaluate the discrete distribution values 
$$
\omega_i = \omega(-L + i \Delta x), \quad i=0,\ldots,n-1, \,\, \Delta x = \tfrac{2L}{n}.
$$
}
\STATE{Set
$$
\boldsymbol{\omega} = \begin{bmatrix} \omega_0 \\ \vdots \\ \omega_{n-1} \end{bmatrix}.
$$
}
\STATE{Evaluate
\begin{equation*}
	\begin{aligned}
C^k &= (\Delta x)^{-1} \left[D \, \mathcal{F}^{-1} \big(\mathcal{F}( D \boldsymbol{\omega} \Delta x)^{\odot k}   \big) \right], \\
\ell_\veps &= \min \{ \ell \in \mathbb{Z} \, : \, -L + \ell \Delta x > \veps \}.
	\end{aligned}
\end{equation*}
}
\STATE{Evaluate the approximation 
$$
\delta( \veps) \approx  \Delta x \sum\nolimits_{\ell=\ell_\veps}^{n-1}  \big(1 - \ee^{\veps - ( - L + \ell \Delta x)} \big) C^k_\ell.
$$
}
\end{algorithmic}
\label{alg:delta}
\end{algorithm}

\subsection{ Computing $\veps(\delta)$ using Newton's method}

In order to get the function $\veps(\delta)$, we compute the inverse of $\delta(\veps)$ using Newton's method.
From \eqref{eq:approx_int} it follows that (see Lemma~\ref{lem:newton} of Appendix) 
\begin{equation} \label{eq:i_repr2}
\delta'(\veps) = - \int_\veps^\infty  \ee^{\veps - s} (\omega *^k \omega ) (s)  \, \dd s.
\end{equation}
Thus, in order to find $\veps$ such that $\delta(\veps) = \bar{\delta}$, we apply Newton's method~\cite{stoer_book} to the function
$\delta(\veps) - \bar{\delta}$ which gives the iteration
$$
\veps_{\ell+1} = \veps_\ell -  \frac{\delta(\veps_\ell) - \bar{\delta}} {\delta'(\veps_\ell) }.
$$
Evaluating $\delta'(\veps)$ for different values of $\veps$ is cheap using the formula \eqref{eq:i_repr2}
and an approximation analogous to \eqref{eq:approx_int}. As is common practice, we use as a stopping criterion
$\abs{\delta({\veps_{\ell}}) - \bar{\delta}} \leq \tau	$
for some prescribed tolerance parameter $\tau$. The iteration was found to converge in all experiments with an initial value $\veps_0=0$.

\subsection{Approximation for varying mechanisms}

Our approach also allows computing privacy cost of a composite mechanism $\mathcal{M}_1 \circ \ldots \circ \mathcal{M}_k$, 
where the PLDs of the mechanisms $\mathcal{M}_{i}$ vary.
This is needed for example when accounting the privacy loss of Stochastic Gradient Langevin Dynamics iterations~\cite{wang15privacy}, where decreasing
the step size increases $\sigma$.

In this case the function $\delta(\veps)$ is given by Thm.~\ref{thm:convolutions} of Appendix by an integral formula of the form
\begin{equation*}
	\begin{aligned}
	\delta( \veps) =  \int_\veps^\infty (1 - \ee^{\veps - s})(\omega_1 * \ldots * \omega_k ) (s)  \, \dd s,
	\end{aligned}
\end{equation*}
where $\omega_i$'s are PLD distributions determined by the mechanisms $\mathcal{M}_i$, $1 \leq i \leq k$.

Denoting $C = (\Delta x)^{-1} \left[D \, \mathcal{F}^{-1} \big(  F_1 \odot \ldots \odot F_k \big) \right]$,
where $F_i = \mathcal{F}( D \boldsymbol{\omega_i} \Delta x )$ and $\boldsymbol{\omega_i}$'s are obtained
from discretisations of $\omega_i$'s (as in Sec.~\ref{subsec:discretisation}), then $\delta(\veps)$ can be approximated
as in \eqref{eq:approx_int}.

\section{Subsampled Gaussian mechanism} \label{Sec:subsampled_gauss}

The main motivation for this work comes from privacy accounting of the subsampled Gaussian mechanism which gives privacy bounds for DP-SGD (see e.g. \cite{Abadi2016}).
In the appendix, we show that the worst case privacy analysis of DP-SGD can be carried out by analysis of one dimensional probability distributions.
We  derive the privacy loss distributions for three different subsampling methods: 
Poisson subsampling with both $\sim_R$- and $\sim_S$-neighbouring relations, sampling without replacement with $\sim_S$-neighbouring relation and
sampling with replacement with $\sim_S$-neighbouring relation.
We note the following related works.
In~\cite{balle2018subsampling} RDP bounds are considered for these three subsampling methods,
in~\cite{wang2019} improved RDP bounds were given for the case of sampling without replacement and
in~\cite{zhu2019} tight RDP bounds were given for the case of Poisson subsampling.

\subsection{Poisson subsampling for $(\varepsilon, \delta, \sim_R)$-DP} \label{subsec:unbounded}

We start with the Poisson subsampling method, where each member of the dataset is included in the stochastic gradient minibatch with probability $q$.
This method is also used in the moments accountant~\cite{Abadi2016}, and also considered in~\cite{meiser2018tight} and~\cite{wang2019}.
As we show in Appendix, the $(\varepsilon, \delta, \sim_R)$-DP analysis of the Poisson subsampling is equivalent to considering the
following one dimensional distributions: 
\begin{equation*}
	\begin{aligned}
		f_X(t) &= q \, \tfrac{1}{\sqrt{2 \pi \sigma^2}} \ee^{ \frac{-(t-1)^2}{2 \sigma^2} }  +
		 (1-q) \tfrac{1}{\sqrt{2 \pi \sigma^2}} \ee^{ \frac{-t^2}{2 \sigma^2}},  \\
 	f_Y(t) &=  \tfrac{1}{\sqrt{2 \pi \sigma^2}} \ee^{ \frac{-t^2}{2 \sigma^2}} .
	\end{aligned}
\end{equation*}
Here $\sigma^2$ denotes the variance of the additive Gaussian noise.
Using Definition~\ref{def:plf}, the privacy loss function is given by
\begin{equation*}
	\begin{aligned}
\mathcal{L}_{X/Y}(t)  &= \log \tfrac{q \, \tfrac{1}{\sqrt{2 \pi \sigma^2}} \ee^{ \frac{-(t-1)^2}{2 \sigma^2} }  +
 (1-q) \tfrac{1}{\sqrt{2 \pi \sigma^2}} \ee^{ \frac{-t^2}{2 \sigma^2} }  }{ \tfrac{1}{\sqrt{2 \pi \sigma^2}} \ee^{ \frac{-t^2}{2 \sigma^2}} } \\
 &= \log \left( q \, \ee^{ \frac{2t-1}{2 \sigma^2} }  + (1-q) \right).
	\end{aligned}
\end{equation*}
Now $\mathcal{L}_{X/Y}(\mathbb{R}) = ( \log(1-q), \infty)$ and
$\mathcal{L}_{X/Y}$ is again a strictly increasing continuously differentiable bijective
function in the whole $\mathbb{R}$. Straightforward calculation shows that
$$
\mathcal{L}_{X/Y}^{-1}(s)  = \sigma^2 \log \frac{\ee^s  - (1-q)}{ q } + \frac{1}{2}.
$$
Moreover,
$$
\frac{\dd}{\dd \, s} \mathcal{L}_{X/Y}^{-1}(s) = \frac{ \sigma^2 \ee^s }{ \ee^s  - (1-q) }.
$$
The privacy loss distribution $\omega_{X/Y}$ is determined by the density function given in Definition~\ref{def:pld}.
Lemma~\ref{lem:dual} and its corollary
explain the observation that generally $\delta_{X/Y} > \delta_{Y/X}$.

\subsection{Sampling without replacement for $(\varepsilon, \delta, \sim_S)$-DP}  \label{subsec:bounded}

We next consider the $\sim_S$-neighbouring relation and sampling without replacement.
In this case the batch size $m$ is fixed and each member of the dataset contributes at most once for each minibatch.
Here $q=m/n$, where $n$ denotes the total number of data samples.
Without loss of generality we consider here the density functions
\begin{equation*}
	\begin{aligned}
		f_X(t) &= q \, \tfrac{1}{\sqrt{2 \pi \sigma^2}} \ee^{ \frac{-(t-1)^2}{2 \sigma^2} }  +
		 (1-q) \tfrac{1}{\sqrt{2 \pi \sigma^2}} \ee^{ \frac{-t^2}{2 \sigma^2}}, \\
 	f_Y(t) &=  q \, \tfrac{1}{\sqrt{2 \pi \sigma^2}} \ee^{ \frac{-(t+1)^2}{2 \sigma^2} }  +
		 (1-q) \tfrac{1}{\sqrt{2 \pi \sigma^2}} \ee^{ \frac{-t^2}{2 \sigma^2}}.
	\end{aligned}
\end{equation*}
The privacy loss function is now given by
\begin{equation*}
	\begin{aligned}
\mathcal{L}_{X/Y}(t)  = \log \Bigg( \frac{ q \, \ee^{ \frac{2t-1}{2 \sigma^2} }  + (1-q) }{q \, \ee^{ \frac{-2t-1}{2 \sigma^2} }  + (1-q) }    \Bigg).
	\end{aligned}
\end{equation*}
We see that $\mathcal{L}_{X/Y}(\mathbb{R}) = \mathbb{R}$ and again that
$\mathcal{L}_{X/Y}$ is a strictly increasing continuously differentiable
function. With a straightforward calculation we find that
\begin{equation*}
	\begin{aligned}
		\mathcal{L}_{X/Y}^{-1}(s) = \sigma^2 \log \Big(   \frac{1}{ 2c  } \big(   -(1-q)(1-\ee^s)  + \sqrt{ (1-q)^2(1-\ee^s)^2 + 4c^2 \ee^s } \big)   \Big),
	\end{aligned}
\end{equation*}
where $c = q \, \ee^{- \frac{1}{2 \sigma^2}}$.

Using Lemma~\ref{lem:dual} and the property
$f_Y(-t) = f_X(t)$, we see that $\delta = \delta_{Y/X} = \delta_{X/Y}$. 

We remark that in $(\varepsilon, \delta, \sim_S)$-DP, the Poisson subsampling with the sampling parameter $\gamma$
is equivalent to the case of the sampling without replacement with $q=\gamma$, as in both cases the differing element is included in the minibatch with probability $\gamma$.

\subsection{Sampling with replacement}  \label{subsec:bounded2}

Consider next the sampling with replacement and the $\sim_S$-neighbouring relation. 
Again the batch size is fixed, however this time each element of the minibatch is drawn from the dataset with probability q. 
Thus the number of contributions of each member of the dataset is not limited.
Then $\ell$, the number of times the differing sample $x'$ is in the batch, is binomially distributed,
i.e., $\ell \sim \mathrm{Binomial}(1/n,m)$, where $m$ denotes the batch size and $n$ the total number of data samples.

Without loss of generality, we consider here the density functions 
\begin{equation*} 
	\begin{aligned}
		f_X(t) &=  \tfrac{1}{\sqrt{2 \pi \sigma^2}}  \sum\limits_{\ell=0}^m \left( \frac{1}{n} \right)^\ell \left( 1- \frac{1}{n} \right)^{m-\ell} {m\choose\ell} \,\ee^{ \frac{-(t-\ell)^2}{2 \sigma^2} }, \\
 	f_Y(t) &=   \tfrac{1}{\sqrt{2 \pi \sigma^2}}  \sum\limits_{\ell=0}^m \left( \frac{1}{n} \right)^\ell \left( 1- \frac{1}{n} \right)^{m-\ell} {m\choose\ell} \,\ee^{ \frac{-(t+\ell)^2}{2 \sigma^2} }.
	\end{aligned}
\end{equation*}
The privacy loss function is then given by
\begin{equation*}
	\begin{aligned}
\mathcal{L}_{X/Y}(t) 
 =  \log \left( \frac{ \sum_{\ell=0}^m  c_\ell x^\ell }{ \sum_{\ell=0}^m  c_\ell x^{-\ell}} \right), 
	\end{aligned}
\end{equation*}
where 
\begin{equation*}
	c_\ell = \left( \frac{1}{n} \right)^\ell \left( 1- \frac{1}{n} \right)^{m - \ell}  {m\choose\ell} \,\ee^{  \frac{ - \ell^2}{2 \sigma^2} }, \quad x = \ee^{  \frac{ t }{ \sigma^2} }.
\end{equation*}
Since $c_\ell>0$ for all $\ell = 1, \ldots, m$, clearly $\sum_{\ell=0}^m  c_\ell x^\ell$ is strictly increasing as a function of $t$ and
$\sum_{\ell=0}^m  c_\ell x^{-\ell}$ is strictly decreasing. Moreover, we see that 
$$
\frac{ \sum_{\ell=0}^m  c_\ell x^\ell }{ \sum_{\ell=0}^m  c_\ell x^{-\ell}} \rightarrow 0 \quad \textrm{as} \,\, t \rightarrow -\infty
$$ 
and  
$$
\frac{ \sum_{\ell=0}^m  c_\ell x^\ell }{ \sum_{\ell=0}^m  c_\ell x^{-\ell}}  \rightarrow \infty \quad \textrm{as} \,\,t \rightarrow \infty.
$$ 
Thus, $\mathcal{L}_{X/Y}(\mathbb{R}) = \mathbb{R}$ and
$\mathcal{L}_{X/Y}(t)$ is a strictly increasing continuously differentiable function in its domain. 
To find $\mathcal{L}_{X/Y}^{-1}(s)$ one needs to solve $\mathcal{L}_{X/Y}(t) = s$,
i.e., one needs to find the single positive real root of a polynomial of order $2m$.
As in the case of subsampling without replacement, here $\delta = \delta_{Y/X} = \delta_{X/Y}$.

\section{Error estimates} \label{sec:err_est}

We give error estimates for the Poisson subsampling method with the neighbouring relation $\sim_R$.
Thus, in this section $\omega$ denotes the PLD density function defined in Sec.~\ref{subsec:unbounded}.
The estimates are determined by the parameters $L$ and $n$,
the truncation interval radius and the number of discretisation points, respectively.

The total error consists of (see Thm.~\ref{thm:total_error} in Appendix)
\begin{enumerate}
	\item The errors arising from the truncation of the convolution integrals and periodisation. 
	\item The error from neglecting the tail integral 
	\begin{equation} \label{eq:tail}
		\int_L^\infty (1 - \ee^{\veps - s})(\omega *^k \omega ) (s)  \, \dd s.
	\end{equation}
	\item The numerical errors in the approximation of the convolutions $ ( \omega \ast^k \omega ) $ 
	and in the Riemann sum approximation \eqref{eq:approx_int}. 
\end{enumerate}

We obtain bounds for the first two sources of error, i.e., for the tail integral \eqref{eq:tail} and the periodisation error, using the Chernoff bound~\cite{wainwright2019}
\begin{equation*} 
\mathbb{P}[ X \geq t] = \mathbb{P}[ \ee^{\lambda X} \geq \ee^{\lambda t} ] \leq \frac{ \mathbb{E}[ \ee^{\lambda X} ] }{\ee^{\lambda t}},
\end{equation*}
which holds for any random variable $X$ and all $\lambda > 0$.
Denoting also the PLD random variable by $\omega$,
the moment generating function $\mathbb{E} [\ee^{\lambda \omega }]$ is related to
the log of the moment generating function of the privacy loss function $\mathcal{L} = \mathcal{L}_{X/Y}$
as follows. Define (see also~\cite{Abadi2016})
$$
\alpha(\lambda) := \log \mathop{\mathbb{E}}_{t \sim f_X(t)} [\ee^{\lambda \mathcal{L}(t)}].
$$
By the change of variable $s = \mathcal{L}(t)$ we have
\begin{equation} \label{eq:pld_lmf}
	\begin{aligned}
		\mathbb{E} [\ee^{\lambda \omega }] &= \int_{-\infty}^\infty \ee^{\lambda s} \omega(s) \, \dd s \\
		&= \int_{ \log(1-q)}^\infty \ee^{\lambda s} f_X( \mathcal{L}^{-1}(s)) \frac{\dd \mathcal{L}^{-1}(s)}{\dd s} \, \dd s \\
		&= \int_{ -\infty}^\infty \ee^{\lambda \mathcal{L}(t)} f_X(t) \, \dd t = \ee^{\alpha(\lambda)}.
	\end{aligned}
\end{equation}
Using existing bounds for $\alpha(\lambda)$ given in~\cite{Abadi2016} and~\cite{mironov2019}, we bound 
$\mathbb{E} [\ee^{\lambda \omega }]$ and obtain the required tail bounds.

\subsection{Periodisation and truncation of convolutions}

We have the following bound for the error arising from the periodisation and the truncation of the convolution integrals.
The proof is given in Appendix, Lemma~\ref{lem:period}.
\begin{lem} \label{lem:first_term}
Let $0<q<\frac{1}{2}$. Let $\omega$ be defined as in Sec.~\ref{subsec:unbounded}, and let $L \geq 1$.
Then, for all $x \in \mathbb{R}$,
\begin{equation*} 
	\begin{aligned}
	&	\abs{\int_\veps^L (\omega \ast^k \omega - \widetilde{\omega} \circledast^k \widetilde{\omega})(x) \, \dd x} \leq 
		L k \sigma  \ee^{ - \frac{ ( \sigma^2  L + C )^2 }{2 \sigma^2} } \\
	& \quad \quad + \ee^{ \alpha(L/2)} \ee^{- \frac{L^2}{2}} + 2 \sum\nolimits_{n=1}^\infty \ee^{k \alpha( nL)} \ee^{-2 (nL)^2},
	\end{aligned}
\end{equation*}
where $C = \sigma^2 \log( \frac{1}{2q}) - \frac{1}{2}$.
\end{lem}
For example, setting $\sigma$, $q$ as in the example of Figure~\ref{fig:bound_lemma3}, and $k=2 \cdot 10^4$,
the first term is $O(10^{-16})$  already for $L=4.0$.
The second term dominates the rest of the bound of Lemma~\ref{lem:first_term} and it is much smaller than
the tail bound \eqref{eq:approx_bound} ($\ee^{ \alpha(L/2)}$ vs. $\ee^{ k \alpha(L/2)}$).
Therefore, this error is much smaller than estimates for the tail integral \eqref{eq:tail} and it is neglected in the numerical estimates.

\subsection{Convolution tail bound}

Let $\omega$ denote the PLD density function. Now, the tail of the integral 
representation for $\delta$ (Thm.~\ref{thm:integral}), with $L>\veps$, can be bounded as
\begin{equation*} 
	\int_L^\infty (1 - \ee^{\veps - s})(\omega *^k \omega ) (s)  \, \dd s < \int_L^\infty (\omega *^k \omega ) (s)  \, \dd s.
\end{equation*}
We consider both upper bounds and estimates for the tail integral of convolutions. 

\subsubsection{Analytic tail bound}

Using the Chernoff bound we derive an analytic bound for the tail integral of convolutions.
In a certain sense this is equivalent to finding bounds for the RDP parameters, since
an RDP bound gives a bound also for the moment generating function $\mathbb{E} [\ee^{\lambda \omega }]$ needed in the Chernoff bound. 
The following result is derived from recent RDP results~\cite{mironov2019}. The proof and an illustration of the result are given in Appendix.
\begin{thm}	\label{thm:a_bound}
Suppose $q\leq \frac{1}{5}$ and $\sigma \geq 4$. Let $L$ be chosen such that $\lambda = L/2$ satisfies
\begin{equation*}
	\begin{aligned}
		1 < & \lambda \leq \frac{1}{2}\sigma^2 c - 2 \log \sigma, \\
		 & \lambda \leq \frac{\frac{1}{2}\sigma^2 c  - \log \, 5 - 2 \log \, \sigma}{c + \log(q \lambda) + 1/(2\sigma^2)},
	\end{aligned}
\end{equation*}
where $c = \log \left( 1 + \frac{1}{q ( \lambda - 1)}   \right)$. Then, we have
$$
\int_L^\infty  ( \omega \ast^k \omega)(s) \, \dd s \leq  \left(  1 + \frac{2q^2(\tfrac{L}{2}+1)\tfrac{L}{2}}{\sigma^2}  \right)^k \ee^{-\frac{L^2}{2}}.
$$
\end{thm}
In order to avoid the restriction on $\sigma$ in Thm.~\ref{thm:a_bound}, we consider an approximative bound.

\subsubsection{Tail bound estimate} \label{subsubsec:num_approx1}

\begin{figure} [h!]
	\begin{center}
  \includegraphics[width=0.5\linewidth]{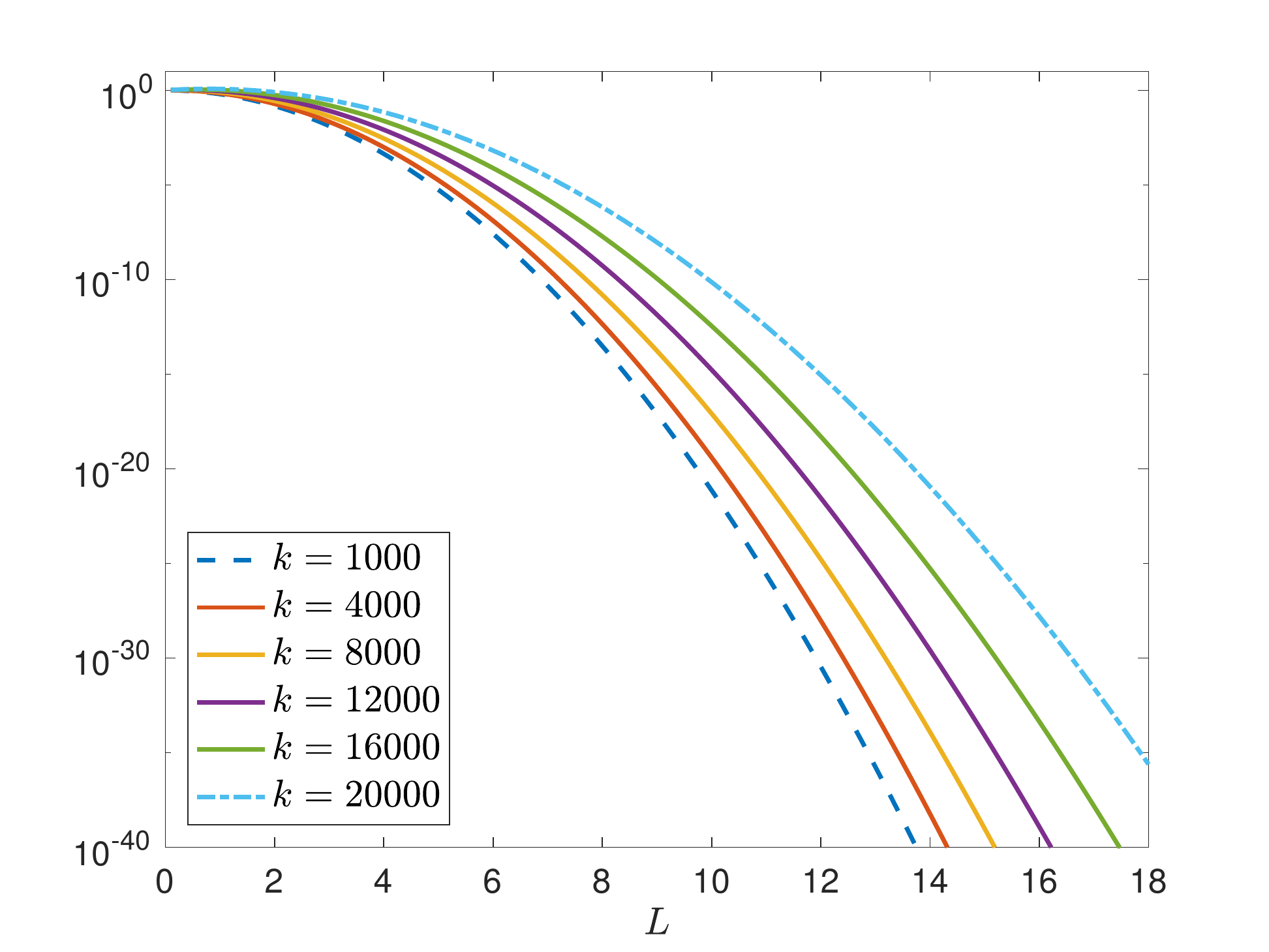}
	\caption{Convergence of the bound \eqref{eq:approx_bound} for $q=0.01$ and $\sigma=2.0$ for different number of compositions $k$. }
 \label{fig:bound_lemma3}
	\end{center}
\end{figure}

We next derive an approximative tail bound using the $\alpha(\lambda)$-bound given in~\cite{Abadi2016}.
Denote $S_k := \sum_{i=1}^k \omega^i$, where $\omega^i$ denotes the PLD random variable of the $i$th mechanism.
Since $\omega^i$'s are independent, $ \mathbb{E} [ \ee^{\lambda S_k}  ]  = \prod_{i=1}^k \mathbb{E} [ \ee^{\lambda \omega^i}  ] $
and the Chernoff bound shows that
\begin{equation*}
	\begin{aligned}
		\int_L^\infty  ( \omega \ast^k \omega)(s) \, \dd s   = \mathbb{P}[ S_k \geq L ] \leq \ee^{k \alpha(\lambda)}  \ee^{- \lambda L} 
	\end{aligned}
\end{equation*}
for any $\lambda > 0$.
We recall the result from~\cite[Lemma 3]{Abadi2016} which holds for the Poisson subsampling method.
\begin{lem} \label{lem:lemma3}
Let $\sigma \geq 1$ and $q < \frac{1}{16 \sigma}$, then for any positive integer $\lambda \leq \sigma^2 \ln \frac{1}{q \sigma}$, 
$$
\alpha(\lambda) \leq \frac{ q^2 \lambda(\lambda+1) }{(1-q) \sigma^2} + \mathcal{O} (q^3\lambda^3/\sigma^3).
$$
\end{lem}
Suppose the conditions of Lemma~\ref{lem:lemma3} hold for $\lambda=L/2$.
Substituting the bound of Lemma~\ref{lem:lemma3} to the Chernoff bound 
and neglecting the $\mathcal{O} (q^3\lambda^3/\sigma^3)$-term  gives the approximative upper bound
\begin{equation} \label{eq:approx_bound}
	\begin{aligned}
   \hspace{-2.19mm}	 \int_L^\infty  ( \omega \ast^k \omega)(s)  \dd s
	 \lessapprox \exp\left({k \frac{ q^2 (\tfrac{L}{2}+1)  \tfrac{L}{2}}{(1-q) \sigma^2} } \right) \ee^{-\tfrac{L^2}{2}}.
	\end{aligned}
\end{equation}
For example, when $q=0.01$ and $\sigma=2.0$, the conditions of Lemma~\ref{lem:lemma3} hold for $\lambda$ up to $\approx 9.5$
(i.e.~\eqref{eq:approx_bound} holds for $L$ up to $\approx 19$).
Figure~\ref{fig:bound_lemma3} shows the convergence of the bound \eqref{eq:approx_bound} in this case.

\subsection{Discretisation errors}

Derivation of discretisation error bounds can be carried out using the so called Euler--Maclaurin formula (Sec.~\ref{subsec:err_exp} in Appendix).
This requires bounds for higher order derivatives of $\omega$.
As an illustrating example,
consider the bound (recall $\Delta x = 2L/n$)
\begin{equation*}
	\begin{aligned}
		\abs{\int_{-L}^L \omega(s) \, \dd s - \Delta x \sum\nolimits_{\ell=0}^{n-1}  \omega(-L + \ell \Delta x)}
	& 	\leq \Delta x \, \omega(L) + \frac{(\Delta x)^2}{12} \max_{t \in [-L,L]} \abs{\omega''(t)} \\
	& 	\leq \Delta x \, \sigma \ee^{ - \frac{ - ( \sigma^2 L + C)^2 }{2 \sigma^2} } + \frac{(\Delta x)^2}{12} \max_{t \in [-L,L]} \abs{\omega''(t)},
	\end{aligned}
\end{equation*}
where $C = \sigma^2 \log( \frac{1}{2q}) - \frac{1}{2}$. 
By Lemma~\ref{lem:boundd}, $\max_t \abs{\omega''(t)}$ has an upper bound $O(\sigma^3/q^3)$.
With bounds for higher order derivatives, tighter error bound could be obtained. 
In a similar fashion, bounds for the errors for the approximation \eqref{eq:approx_int}
could be derived. 
However, we resort to numerical estimates. 

\subsubsection{Estimate for the discretisation error}  \label{subsec:num_approx2}

Consider the error arising from the Riemann sum
$$
I_n := \Delta x \sum\nolimits_{\ell=\ell_\veps}^{n-1}  \big(1 - \ee^{\veps - ( - L + \ell \Delta x)} \big) C^k_\ell.
$$
As we show in Sec.~\ref{subsec:err_exp} of Appendix, it holds
\begin{equation*}
	\begin{aligned}
 	E_n := & \int_\veps^L (1 - \ee^{\veps - s})(\widetilde{\omega} \circledast^k \widetilde{\omega} ) (s)  \, \dd s  - I_n \\
 = & K \Delta x + O((\Delta x)^2) = K  \frac{2L}{n} + O\Big(\Big(\frac{2L}{n}\Big)^2\Big)
	\end{aligned}
\end{equation*}
for some constant $K$ independent of $n$. 
Therefore,
$$
2 (I_n - I_{2n}) = E_n + O((\Delta x)^2)
$$
which leads us to use as an estimate
\begin{equation} \label{eq:errest}
	\mathrm{err}(L,n):= 2 \abs{I_n - I_{2n}}
\end{equation}
for the numerical error $E_n$.

\section{Experiments}

In all experiments, we consider the Poisson subsampling with $(\varepsilon, \delta, \sim_R)$-DP (Sec.~\ref{subsec:unbounded}).

We first illustrate the numerical convergence of FA for $\delta(\veps)$
and the estimates \eqref{eq:approx_bound} and \eqref{eq:errest}, when $k=10^4$, $q=0.01$, $\sigma=1.5$ and $\veps=1.0$
(Tables~\ref{table:conv1} and~\ref{table:conv2}). 
We emphasise that the error estimates~\eqref{eq:approx_bound} and~\eqref{eq:errest} represent the distance
to the tight $\delta(\veps)$-value. This indicates that the approximations converge to the actual tight $\delta(\veps)$-values.

\begin{table}[h!]
\begin{center}
\begin{tabular}{rcc}
 \hline
 $n \quad $ &   FA & $\mathrm{err}(L,n)$  \\
 \hline
 $5 \cdot 10^4$  &  0.0491228786423 &  $2.01 \cdot 10^{-2} $  \\
 $1 \cdot 10^5$  &  0.0496089458356 &  $3.12 \cdot 10^{-4} $  \\
 $2 \cdot 10^5$  &  0.0496013846114 &  $1.06 \cdot 10^{-6} $  \\
 $4 \cdot 10^5$  &  0.0496014103882 &  $1.71 \cdot 10^{-9} $  \\
 $8 \cdot 10^5$  &  0.0496014103252 &  $2.66 \cdot 10^{-11} $  \\
 $1.6 \cdot 10^6$  &  0.0496014103146 & $ 8.88 \cdot 10^{-12} $  \\
 $3.2 \cdot 10^6$  &  0.0496014103163 &  $2.22 \cdot 10^{-12} $  \\
 \hline
 \end{tabular}
 \bigskip
 \caption{Convergence of $\delta(\veps)$-approximation with respect to $n$ (when $L= 12$) and the estimate~\eqref{eq:errest}.
	 The tail bound estimate \eqref{eq:approx_bound} is $O(10^{-24})$.}
 \label{table:conv1}
\end{center}
\end{table}

\begin{table}[h!]
\begin{center}
\begin{tabular}{rccc}
 \hline
 $L $ &   FA & estimate \eqref{eq:approx_bound} \\
 \hline
 $2.0$  &  0.0422160172923 & $ 3.32 \cdot 10^{-1} $  \\
 $4.0$  &  0.0496008932869 & $ 4.96 \cdot 10^{-3} $  \\
 $6.0$  &  0.0496014103158 & $ 3.32 \cdot 10^{-6} $  \\
 $8.0$  &  0.0496014103134 & $ 1.00 \cdot 10^{-10} $  \\
 $10.0$  &  0.0496014103134 & $ 1.36 \cdot 10^{-16} $  \\
 $12.0$  &  0.0496014103163 & $ 8.30 \cdot 10^{-24} $  \\
 \hline
 \end{tabular}
 \bigskip
 \caption{Convergence of the $\delta(\veps)$-approximation with respect to $L$ (when $n=3.2 \cdot 10^6$) 
 and the error estimate~\eqref{eq:approx_bound}.
	  The estimate $\mathrm{err}(L,n) = O(10^{-12})$. }
 \label{table:conv2}
\end{center}
\end{table}

\begin{figure} [ht!]
\centering
\begin{subfigure}{.475\textwidth}
  \centering
  \includegraphics[width=.89\linewidth]{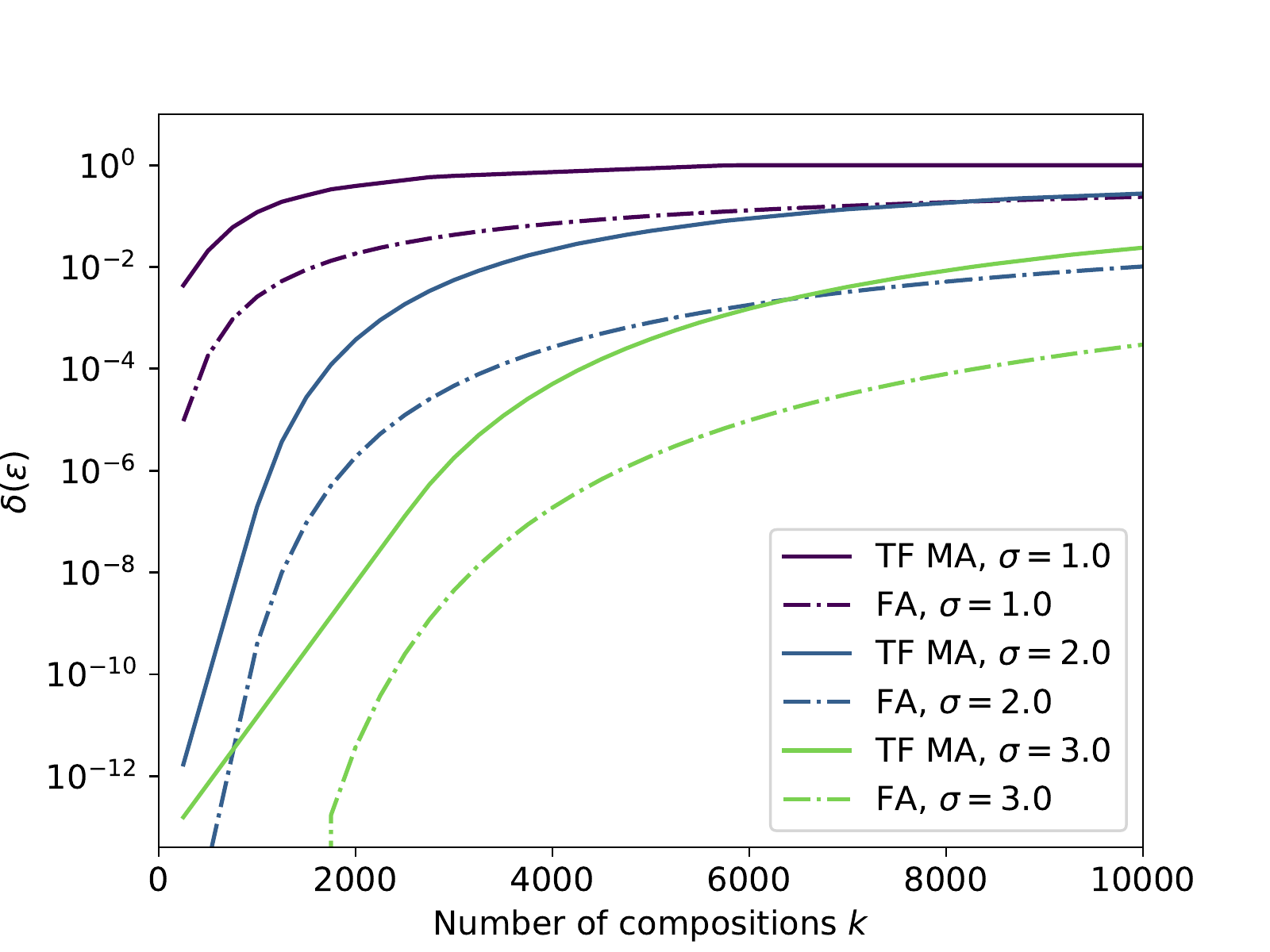}
		\caption{
		$\delta(\veps)$ as a function of $k$ for $\veps = 1.0$. }
\end{subfigure}
~
\begin{subfigure}{.475\textwidth}
  \centering
  \includegraphics[width=.89\linewidth]{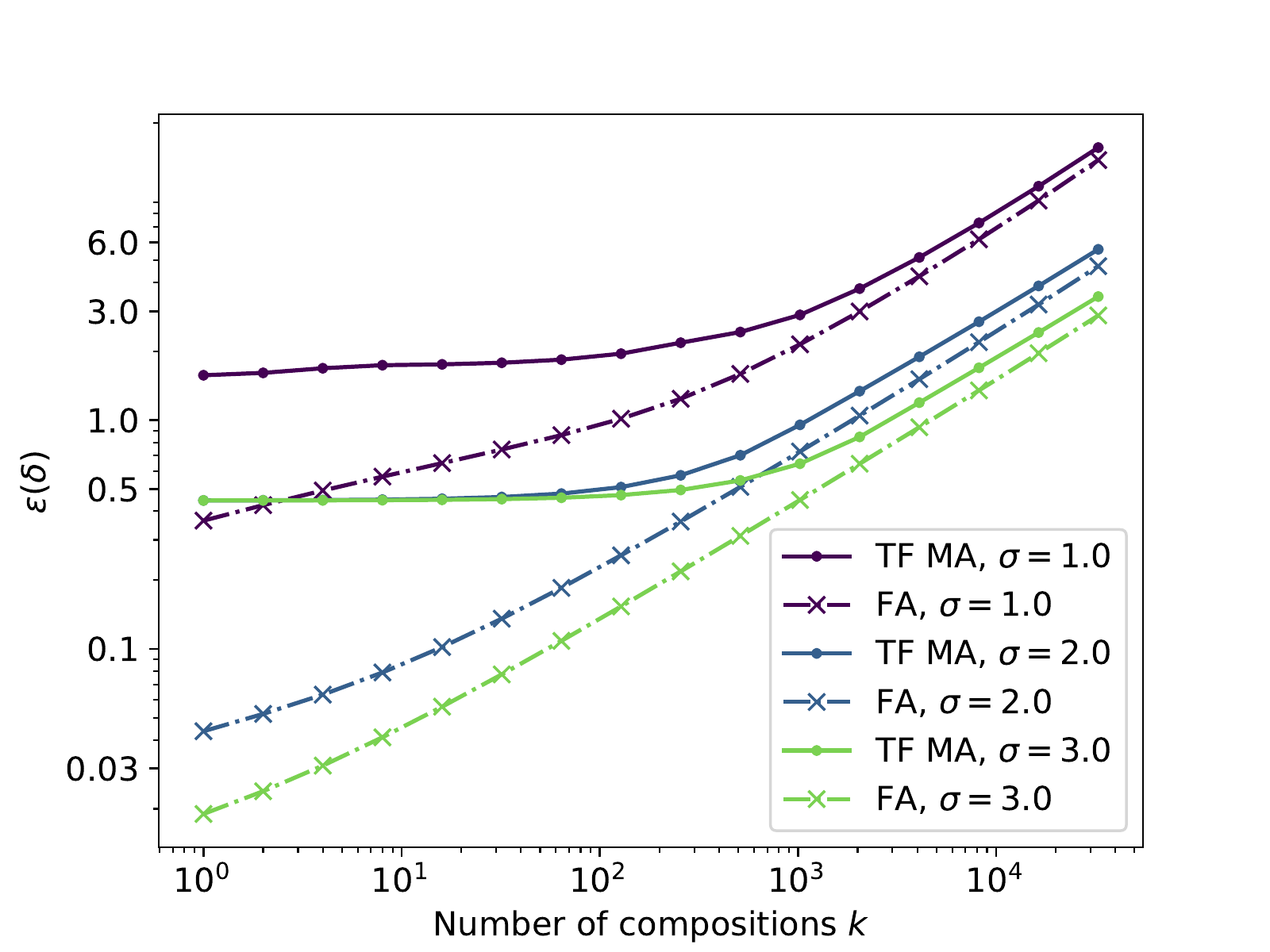}
	\caption{
	$\veps(\delta)$ as a function of $k$ for $\delta = 10^{-6}$. }		
		\label{fig:test_R1}
\end{subfigure}   
\caption{
Comparison of the Tensorflow moments accountant and the Fourier accountant. Here $q=0.01$.}
\label{fig:tf}
\end{figure}

We next compare the Fourier accountant method to the privacy accountant method included in the Tensorflow library~\cite{tensorflow}
which is the moments accountant method~\cite{Abadi2016} (Figure~\ref{fig:tf}). 
We use $q=0.01$ and $\sigma \in \{1.0,2.0,3.0\}$, for number of compositions $k$ up to $10^4$.
We set the parameters $L=12$ and $n=5 \cdot 10^6$ for the approximation of the exact integral.
Then, for $\sigma=1.0$, the tail integral error estimate \eqref{eq:approx_bound} is at most $O(10^{-13})$
and the estimate $\mathrm{err}(L,n)$ is at most $O(10^{-10})$. For $\sigma=2.0,3.0$ the error estimates are smaller.

We next compare FA to the RDP accountant method described in~\cite{zhu2019} (Figure~\ref{fig:poisson_rdp}). 
Although the RDP accountant gives tight RDP-bounds, there is a small gap to the tight 
$(\varepsilon, \delta, \sim_R)$-DP. 

As we see from Figures~\ref{fig:test_R1} and~\ref{fig:poisson_rdp}, the moments accountant and the RDP bound of~\cite{zhu2019} do not
capture the true $\veps$-bound for small number of compositions $k$, whereas FA gives tight bounds also in this case.

\begin{figure} [h!]
  \centering
  \includegraphics[width=.5\linewidth]{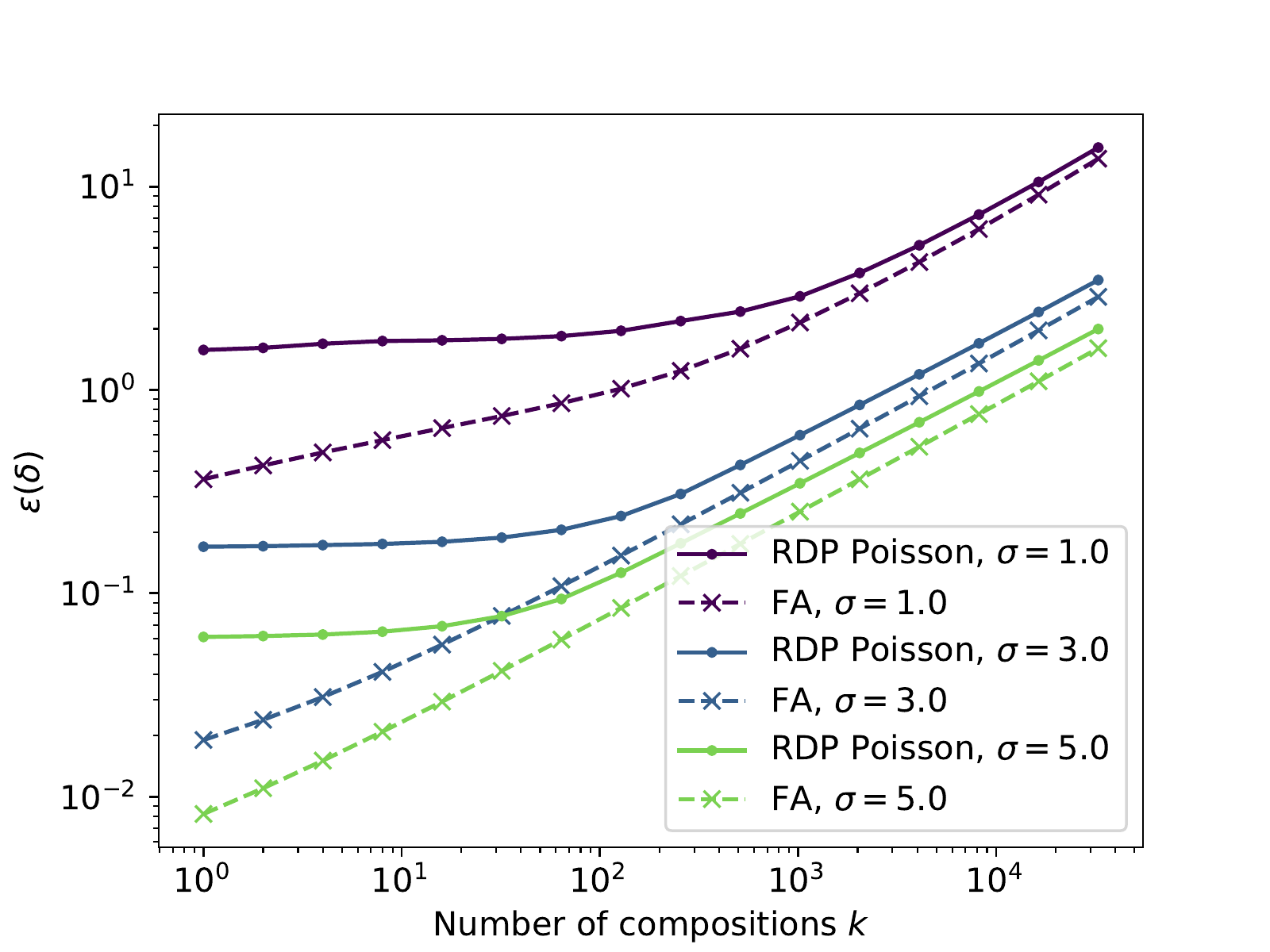}
		\label{fig:poisson_rdp}
	\caption{Comparison of the RDP bound
	for the Poisson subsampling~\cite{zhu2019} and FA. 
	Here $\delta=10^{-6}$, $q=0.01$. }
	\label{fig:poisson_rdp}
\end{figure}

\begin{figure} [h!]
  \centering
  \includegraphics[width=.5\linewidth]{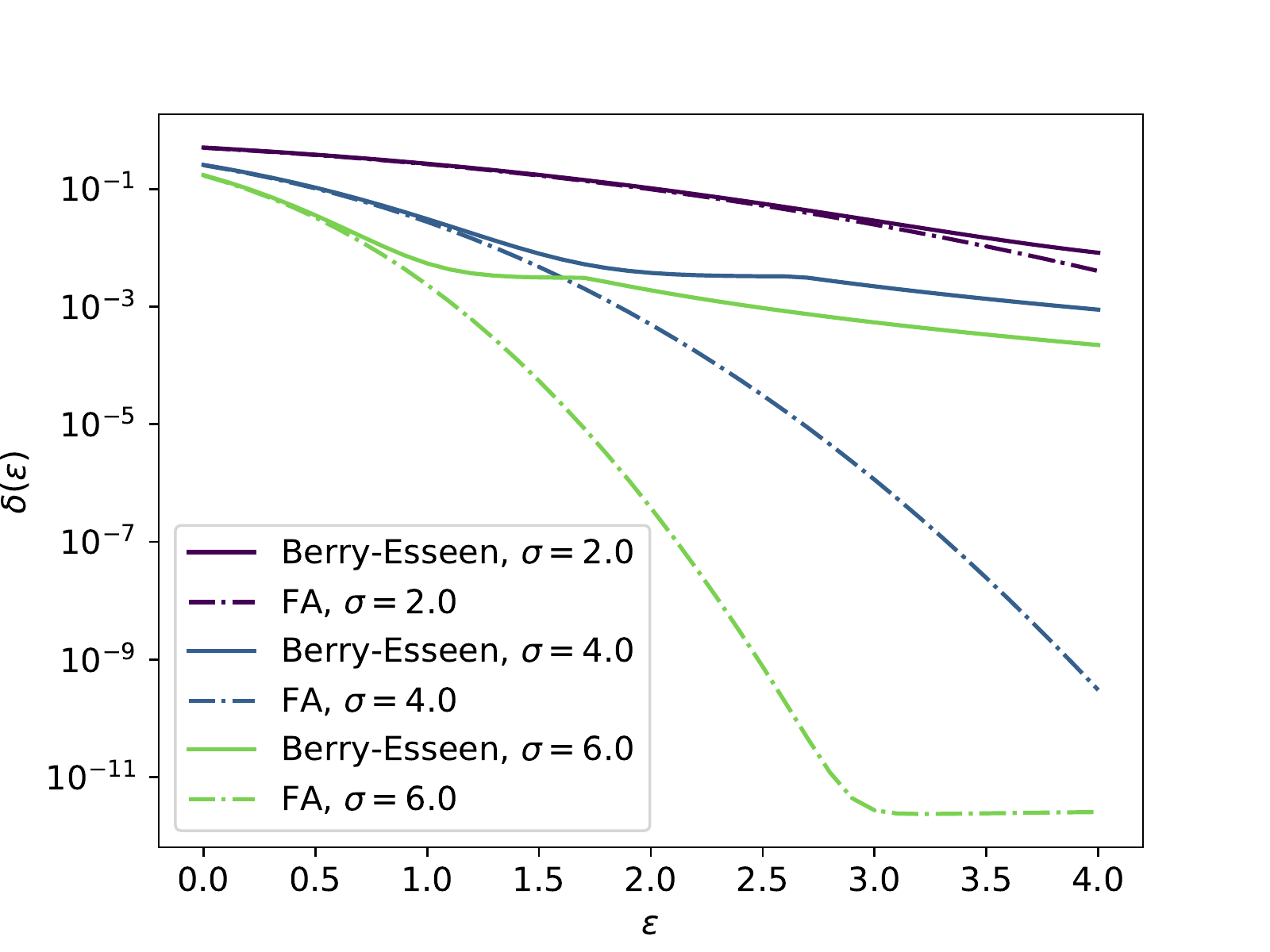}
\caption{Comparison of the Berry--Esseen bound 
and FA for $(\varepsilon, \delta, \sim_R)$-DP. Here $k=5 \cdot 10^4$, $q=0.01$.}
\label{fig:be}
\end{figure}

Figure~\ref{fig:be} shows a comparison of FA to the Berry--Esseen theorem based bound 
given in~\cite[Thm.\;6]{sommer2019privacy}. The Berry--Esseen bound suffers from the 
error term which converges $O(k^{-\frac{1}{2}})$.

\begin{figure} [h!]
  \centering
  \includegraphics[width=.5\linewidth]{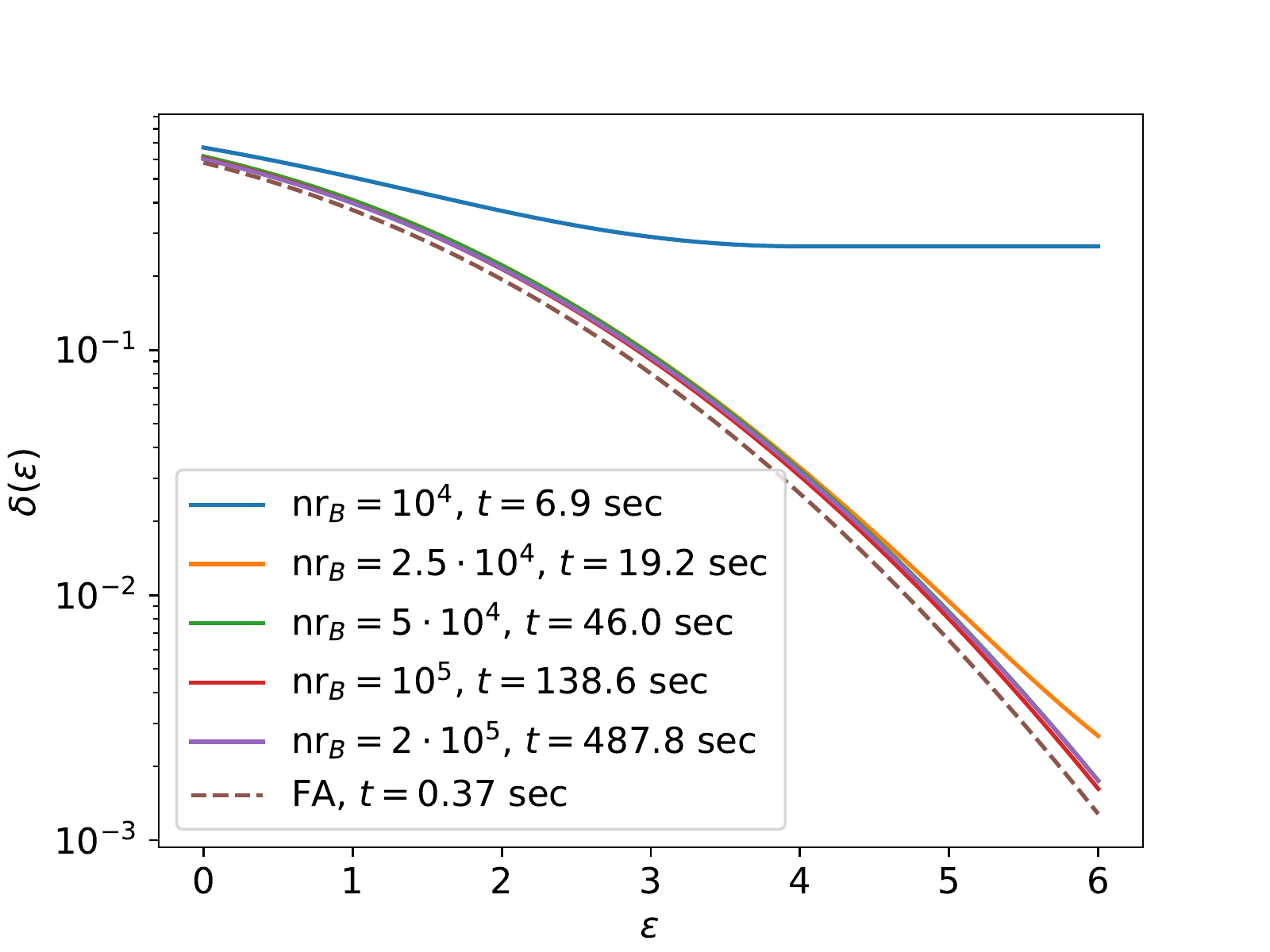}
	\caption{
	Comparison of the Privacy Buckets algorithm ($n_B=$ number of buckets) and FA. 
	Legend contains compute times. Here $k=2^{12}$, $\sigma=1.0$, $q=0.02$.}
\label{fig:buckets}
\end{figure}

Lastly, we compare FA to the Privacy Buckets (PB) algorithm described in~\cite{meiser2018tight} (see~Figure~\ref{fig:buckets}).
The additional ratio parameter of PB was tuned for the experiments.
The algorithm seems to suffer from some instabilities which is also mentioned in~\cite{meiser2018tight}.
For larger $\sigma$ and smaller $q$ PB gave bounds closer to that of FA, however the compute times were always
much bigger, as in experiments of~Figure~\ref{fig:buckets}.

\section{Conclusions}

We have presented a novel approach for computing tight privacy bounds for DP.
Although we have focused on the subsampled Gaussian mechanism (with various subsampling strategies), our method is applicable to any
continuous mechanism satisfying the assumptions of Definition~\ref{def:plf}. Using the existing RDP bounds we were able to give analytical bounds to some central
quantities in the error analysis. For the errors arising from the numerical integration, we gave numerical estimates.
As future work, it would be interesting
to carry out a full error analysis for the discretisation error and also to implement the method on other mechanisms
than the subsampled Gaussian mechanism. Moreover, evaluating for compositions involving both continuous and discrete valued mechanisms would 
also be an interesting objective.

%

\bibliographystyle{plain}
\bibliography{pld}

\begin{thebibliography}{10}

\bibitem{tensorflow}
Mart{\'\i}n Abadi, Paul Barham, Jianmin Chen, Zhifeng Chen, Andy Davis, Jeffrey
  Dean, Matthieu Devin, Sanjay Ghemawat, Geoffrey Irving, Michael Isard, et~al.
\newblock Tensorflow: A system for large-scale machine learning.
\newblock In {\em 12th USENIX Symposium on Operating Systems Design and
  Implementation (OSDI 16)}, pages 265--283, 2016.

\bibitem{Abadi2016}
Martin Abadi, Andy Chu, Ian Goodfellow, H.~Brendan McMahan, Ilya Mironov, Kunal
  Talwar, and Li~Zhang.
\newblock Deep learning with differential privacy.
\newblock In {\em Proc. CCS 2016}, 2016.

\bibitem{balle2018subsampling}
Borja Balle, Gilles Barthe, and Marco Gaboardi.
\newblock Privacy amplification by subsampling: Tight analyses via couplings
  and divergences.
\newblock In {\em Advances in Neural Information Processing Systems}, pages
  6277--6287, 2018.

\bibitem{Bassily2014}
Raef Bassily, Adam Smith, and Abhradeep Thakurta.
\newblock Private empirical risk minimization: Efficient algorithms and tight
  error bounds.
\newblock In {\em Proceedings of the 2014 IEEE 55th Annual Symposium on
  Foundations of Computer Science}, FOCS '14, pages 464--473, Washington, DC,
  USA, 2014. IEEE Computer Society.

\bibitem{Beimel2013}
Amos Beimel, Kobbi Nissim, and Uri Stemmer.
\newblock Characterizing the sample complexity of private learners.
\newblock In {\em Proceedings of the 4th Conference on Innovations in
  Theoretical Computer Science}, ITCS '13, pages 97--110, New York, NY, USA,
  2013. ACM.

\bibitem{chaudhuri06}
Kamalika Chaudhuri and Nina Mishra.
\newblock When random sampling preserves privacy.
\newblock In Cynthia Dwork, editor, {\em Advances in Cryptology - CRYPTO 2006},
  pages 198--213, Berlin, Heidelberg, 2006. Springer Berlin Heidelberg.

\bibitem{Chaudhuri2011}
Kamalika Chaudhuri, Claire Monteleoni, and Anand~D. Sarwate.
\newblock Differentially private empirical risk minimization.
\newblock {\em J. Mach. Learn. Res.}, 12:1069--1109, July 2011.

\bibitem{cooley1965}
James~W Cooley and John~W Tukey.
\newblock An algorithm for the machine calculation of complex fourier series.
\newblock {\em Mathematics of computation}, 19(90):297--301, 1965.

\bibitem{dwork_et_al_2006}
Cynthia Dwork, Frank McSherry, Kobbi Nissim, and Adam Smith.
\newblock Calibrating noise to sensitivity in private data analysis.
\newblock In {\em Proc. TCC 2006}, pages 265--284. 2006.

\bibitem{DworkRoth}
Cynthia Dwork and Aaron Roth.
\newblock The algorithmic foundations of differential privacy.
\newblock {\em Found. Trends Theor. Comput. Sci.}, 9(3--4):211--407, August
  2014.

\bibitem{dwork2010}
Cynthia Dwork, Guy~N. Rothblum, and Salil Vadhan.
\newblock Boosting and differential privacy.
\newblock In {\em Proceedings of the 2010 IEEE 51st Annual Symposium on
  Foundations of Computer Science}, FOCS '10, pages 51--60, Washington, DC,
  USA, 2010. IEEE Computer Society.

\bibitem{kairouz2017}
Peter Kairouz, Sewoong Oh, and Pramod Viswanath.
\newblock The composition theorem for differential privacy.
\newblock {\em IEEE Transactions on Information Theory}, 63(6):4037--4049, June
  2017.

\bibitem{meiser2018tight}
Sebastian Meiser and Esfandiar Mohammadi.
\newblock Tight on budget?: Tight bounds for r-fold approximate differential
  privacy.
\newblock In {\em Proceedings of the 2018 ACM SIGSAC Conference on Computer and
  Communications Security}, pages 247--264. ACM, 2018.

\bibitem{mironov2017}
Ilya Mironov.
\newblock R\'enyi differential privacy.
\newblock In {\em 2017 IEEE 30th Computer Security Foundations Symposium
  (CSF)}, pages 263--275, Aug 2017.

\bibitem{mironov2019}
Ilya Mironov, Kunal Talwar, and Li~Zhang.
\newblock R\'enyi differential privacy of the sampled gaussian mechanism.
\newblock {\em arXiv preprint arXiv:1908.10530}, 2019.

\bibitem{Rajkumar2012}
Arun Rajkumar and Shivani Agarwal.
\newblock A differentially private stochastic gradient descent algorithm for
  multiparty classification.
\newblock In {\em Proc. AISTATS 2012}, pages 933--941, 21--23 Apr 2012.

\bibitem{sommer2019privacy}
David~M Sommer, Sebastian Meiser, and Esfandiar Mohammadi.
\newblock Privacy loss classes: The central limit theorem in differential
  privacy.
\newblock {\em Proceedings on Privacy Enhancing Technologies},
  2019(2):245--269, 2019.

\bibitem{Song2013}
Shuang Song, Kamalika Chaudhuri, and Anand~D. Sarwate.
\newblock Stochastic gradient descent with differentially private updates.
\newblock In {\em Proc. GlobalSIP 2013}, pages 245--248, 2013.

\bibitem{stockham1966}
Thomas~G Stockham~Jr.
\newblock High-speed convolution and correlation.
\newblock In {\em Proceedings of the April 26-28, 1966, Spring joint computer
  conference}, pages 229--233. ACM, 1966.

\bibitem{stoer_book}
Josef Stoer and Roland Bulirsch.
\newblock {\em Introduction to numerical analysis}, volume~12.
\newblock Springer Science \& Business Media, 2013.

\bibitem{wainwright2019}
Martin~J Wainwright.
\newblock {\em High-dimensional statistics: A non-asymptotic viewpoint},
  volume~48.
\newblock Cambridge University Press, 2019.

\bibitem{wang2019}
Yu-Xiang Wang, Borja Balle, and Shiva Kasiviswanathan.
\newblock Subsampled {R}\'enyi differential privacy and analytical moments
  accountant.
\newblock In {\em Proc. AISTATS 2019}, 2019.

\bibitem{wang15privacy}
Yu{-}Xiang Wang, Stephen~E. Fienberg, and Alexander~J. Smola.
\newblock Privacy for free: Posterior sampling and stochastic gradient {M}onte
  {C}arlo.
\newblock In {\em Proc. ICML 2015}, pages 2493--2502, 2015.

\bibitem{zhu2019}
Yuqing Zhu and Yu-Xiang Wang.
\newblock Poission subsampled {R}{\'e}nyi differential privacy.
\newblock In {\em International Conference on Machine Learning}, pages
  7634--7642, 2019.

\end{thebibliography}

\newpage

\appendix 
\section{Proofs for the results of Section~\ref{sec:pld}}

\subsection{Integral representation for exact DP-guarantees}

Throughout this section we denote for neighbouring datasets
$X$ and $Y$ the density function of $\mathcal{M}(X)$ with $f_X(t)$ and the density function of $\mathcal{M}(Y)$ with $f_Y(t)$.

\begin{defn} \label{dfn:tightADP}	
A randomised algorithm $\mathcal{M}$ with an output of continuous one dimensional distributions satisfies 
$(\varepsilon,\delta)$-DP if for every set $S \subset \mathbb{R}$
and every neighbouring datasets $X$ and $Y$
$$
\int_S f_X(t) \, \dd t \leq \ee^\varepsilon \int_S f_Y(t) \, \dd t  + \delta \quad \textrm{and} \quad
\int_S f_Y(t) \, \dd t \leq \ee^\varepsilon \int_S f_X(t) \, \dd t  + \delta.
$$
We call $\mathcal{M}$ tightly $(\varepsilon,\delta)$-DP, if there does not exist $\delta' < \delta$
such that $\mathcal{M}$ is $(\varepsilon,\delta')$-DP.
\end{defn}
The following auxiliary lemma is needed to obtain the representation given by Lemma \ref{lem:maxrepr}
(see~\cite[Lemma 1]{meiser2018tight} for the discrete valued version of the result).
\begin{lem} \label{lem:tight_d}
$\mathcal{M}$ is tightly $(\veps,\delta)$-DP with
\begin{equation} \label{eq:max_eq}
\delta(\veps) = \max_{X \sim Y} \Bigg\{ \int\limits_\mathbb{R}  \max \{  f_X(t) - \ee^\veps f_Y(t) ,0  \} \, \dd t,  
\int\limits_\mathbb{R}  \max \{  f_Y(t) - \ee^\veps f_X(t) ,0  \} \, \dd t \Bigg\}.
\end{equation}
\begin{proof}
Assume $\mathcal{M}$ is tightly $(\veps,\delta)$-DP. Then, for every set $S \subset \mathbb{R}$ and every neighbouring datasets $X$ and $Y$,
\begin{equation*}
	\begin{aligned}
		\int\limits_S   f_X(t) - \ee^\veps f_Y(t) \, \dd t
		\leq & \int\limits_S  \max \{  f_X(t) - \ee^\veps f_Y(t) ,0  \} \, \dd t \\
		\leq & \int\limits_\mathbb{R}  \max \{  f_X(t) - \ee^\veps f_Y(t) ,0  \} \, \dd t.
	\end{aligned}
\end{equation*}	
We get an analogous bound for $\int_S f_Y(t) - \ee^\veps f_X(t) \, \dd t$.
By Definition~\ref{dfn:tightADP},
$$
\delta \leq \max \Bigg\{ \int\limits_\mathbb{R}  \max \{  f_X(t) - \ee^\veps f_Y(t) ,0  \} \, \dd t,  
\int\limits_\mathbb{R}  \max \{  f_Y(t) - \ee^\veps f_X(t) ,0  \} \, \dd t \Bigg\}.
$$
To show that the above inequality is tight, consider the set
$$
S = \{ t \in \mathbb{R} \, : \, f_X(t) \geq \ee^\veps f_Y(t) \}.
$$
Then, 
\begin{equation} \label{eq:caseAB}
	\begin{aligned}
		\int\limits_S  f_X(t) - \ee^\veps f_Y(t) \, \dd t &= \int\limits_S  \max \{ f_X(t) - \ee^\veps f_Y(t),0 \} \, \dd t \\
		&= \int\limits_\mathbb{R}  \max \{ f_X(t) - \ee^\veps f_Y(t),0 \} \, \dd t.
	\end{aligned}
\end{equation}
Next, consider the set
$$
S = \{ t \in \mathbb{R} \, : \, f_Y(t) \geq \ee^\veps f_X(t) \}.
$$
Similarly,
\begin{equation} \label{eq:caseBA}
	\begin{aligned}
		\int\limits_S  f_Y(t) - \ee^\veps f_{X}(t) \, \dd t = \int\limits_\mathbb{R}  \max \{ f_Y(t) - \ee^\veps f_X(t),0 \} \, \dd t.
	\end{aligned}
\end{equation}
From \eqref{eq:caseAB} and \eqref{eq:caseBA} it follows that there exists a set $S \subset \mathbb{R}$ such that either
$$
\int_S f_X(t) \, \dd t = \ee^\varepsilon \int_S f_Y(t) \, \dd t  + \delta \quad \textrm{or} \quad
\int_S f_Y(t) \, \dd t = \ee^\varepsilon \int_S f_X(t) \, \dd t  + \delta
$$
for $\delta$ given by \eqref{eq:max_eq}. This shows that $\delta$ given by \eqref{eq:max_eq} is tight.
\end{proof}
\end{lem}
The next lemma gives an integral representation for the right hand side of \eqref{eq:max_eq}
involving the distribution function of the PLD (see also Lemma 5 of \cite{sommer2019privacy}).
First we need the following definition.
 
\begin{defn} \label{def:plf}
Let $\mathcal{M} \, : \, \mathcal{X}^N \rightarrow \mathbb{R}$ be a randomised mechanism and
let $X \sim Y$. Let $f_X(t)$ denote the density function of
$\mathcal{M}(X)$ and $f_Y(t)$ the density function of $\mathcal{M}(Y)$. Assume
$f_X(t)>0$ and $f_Y(t) > 0$ for all $t \in \mathbb{R}$.
 We define the privacy loss function
of $f_X$ over $f_Y$ as
$$
\mathcal{L}_{X/Y}(t)  = \log \frac{f_X(t)}{f_Y(t)}.
$$
\end{defn}

\begin{lem}  \label{lem:maxrepr}
Let $\mathcal{M}$ be defined as above. 
$\mathcal{M}$  is tightly $(\varepsilon,\delta)$-DP for 
$$
\delta(\veps) = \max_{X \sim Y} \max \{ \delta_{X/Y}(\veps), \delta_{Y/X}(\veps) \}, 
$$
where
\begin{equation*}
	\begin{aligned}
		\delta_{X/Y}(\veps) &= \int\limits_{\mathcal{L}_{X/Y}(\mathbb{R}) \, \cap \, [\veps,\infty)  }  
		(1-\ee^{\veps - s})  f_X\Big( \mathcal{L}^{-1}_{X/Y}(s) \Big) \frac{\dd \mathcal{L}^{-1}_{X/Y}(s)}{\dd s}   \, \dd s, \\
		\delta_{Y/X}(\veps) &= \int\limits_{\mathcal{L}_{Y/X}(\mathbb{R}) \, \cap \, [\veps,\infty)  }   (1-\ee^{\veps - s})  f_Y\Big( \mathcal{L}^{-1}_{Y/X}(s) \Big) \frac{\dd \mathcal{L}^{-1}_{Y/X}(s)}{\dd s}   \, \dd s. 
	\end{aligned}
\end{equation*}
\begin{proof}
Consider the privacy loss function $\mathcal{L}_{X/Y}(t) = \log \frac{f_X(t)}{f_Y(t)}$. Denote $s =\mathcal{L}_{X/Y}(t)$.
Then, it clearly holds $f_Y(t) = \ee^{-s } f_X(t)$ and
\begin{equation} \label{eq:cases}
	\begin{aligned}
	  \max \{  f_X(t) - \ee^\veps f_Y(t) ,0  \} = & \max \{ 0, (1-\ee^{\veps - s}) f_X(t)  \} \\
		 = & \begin{cases}
			(1-\ee^{\veps - s}) f_X(t), &\text{ if } s > \veps,\\
				0, &\text{ otherwise.}
		\end{cases}
	\end{aligned}
\end{equation}
Consider next the integral $\int_\mathbb{R}  \max \{  0, f_X(t) - \ee^\veps f_Y(t)  \} \, \dd t$.
By making the change of variables $t = \mathcal{L}^{-1}_{X/Y}(s)$
and using \eqref{eq:cases}, we see that
\begin{equation*} 
	\begin{aligned}
		\int\limits_\mathbb{R}  \max \{  0, f_X(t) - \ee^\veps f_Y(t)  \} \, \dd t 
		& = \int\limits_\mathbb{R}  \max \{ 0, (1-\ee^{\veps - s}) f_X(t)  \} \, \dd t \\
		& = \int\limits_{\mathcal{L}_{X/Y}(\mathbb{R})}  \max \bigg\{ 0, \, \big(1-\ee^{\veps - s} \big) 
		f_X\big( \mathcal{L}^{-1}_{X/Y}(s) \big) \frac{\dd \mathcal{L}^{-1}_{X/Y}(s)}{\dd s}  \bigg\} \, \dd s \\ 
		& =  \int\limits_{\mathcal{L}_{X/Y}(\mathbb{R}) \, \cap \, [\veps,\infty)  } 
		  (1-\ee^{\veps - s})  f_X\big( \mathcal{L}^{-1}_{X/Y}(s) \big) \frac{\dd \mathcal{L}^{-1}_{X/Y}(s)}{\dd s} \, \dd s,
	\end{aligned}
\end{equation*}
since $\tfrac{\dd \mathcal{L}^{-1}_{X/Y}(s)}{\dd s} \geq 0$ for all $s \in \mathcal{L}_{X/Y}(\mathbb{R})$. 
Analogously, we see that
\begin{equation*} 
\int\limits_\mathbb{R}  \max \{  0, f_Y(t) - \ee^\veps f_X(t)  \} \, \dd t = 
\int\limits_{\mathcal{L}_{Y/X}(\mathbb{R}) \, \cap \, [\veps,\infty)  }  
(1-\ee^{\veps - s})  f_Y\big( \mathcal{L}^{-1}_{Y/X}(s) \big) \frac{\dd \mathcal{L}^{-1}_{Y/X}(s)}{\dd s} \, \dd s.	
\end{equation*}
The claim follows then from Lemma~\ref{lem:tight_d}. 
\end{proof}
\end{lem}
\begin{defn} \label{def:pld}
	Let the assumptions of Definition~\ref{def:plf} of the main text hold and
	 suppose $\mathcal{L}_{X/Y} \, : \,\mathbb{R} \rightarrow D$, $D \subset \mathbb{R}$ is a continuously differentiable bijective function.
 The privacy loss distribution (PLD) of $\mathcal{M}(X)$ over $\mathcal{M}(Y)$ is defined to be a random variable
which has the density function 
\begin{equation*}
	\omega_{X/Y}(s)= \begin{cases}
		 f_X\big( \mathcal{L}_{X/Y}^{-1}(s)  \big)  \,  \frac{\dd \mathcal{L}_{X/Y}^{-1}(s)}{\dd s}, & s \in \mathcal{L}_{X/Y}(\mathbb{R}),\\
		 0, &\text{else.}
	\end{cases}
\end{equation*}
\end{defn}
We directly get from Lemma~\ref{lem:maxrepr} the following representation.
\begin{cor} \label{cor:delta}
A randomised algorithm $\mathcal{M}$ with an output of continuous one dimensional distributions is tightly $(\varepsilon,\delta)$-DP for
\begin{equation} \label{eq:pldrepr2}
\delta(\veps) = \max_{X \sim Y} \max \{ \delta_{X/Y}(\veps), \delta_{Y/X}(\veps) \}, 
\end{equation}
where
\begin{equation*} 
	\begin{aligned}
		\delta_{X/Y}(\veps) = \int\limits_\veps^\infty  (1-\ee^{\veps - s}) \, \omega_{X/Y}(s)  \, \dd s,  \quad
		\delta_{Y/X}(\veps) = \int\limits_\veps^\infty  (1-\ee^{\veps - s}) \, \omega_{Y/X}(s)   \, \dd s. 
	\end{aligned}
\end{equation*}
\end{cor}

\subsection{Privacy loss distribution of compositions}

In order to use the representation given by Corollary \ref{cor:delta} for a composition of
several mechanisms, we need to be able to evaluate the privacy loss distribution for compositions.
This is given in the following theorem which is a continuous version of~\cite[Thm.\;1]{sommer2019privacy}.

\begin{thm} \label{thm:convolutions} 
Let $X,Y$ be adjacent
datasets and let $f_X(t)$ denote the density function of $\mathcal{M}(X)$, $f_Y(t)$ that of $\mathcal{M}(Y)$,
$f_{X'}(t)$ that of $\mathcal{M'}(X)$ and $f_{Y'}(t)$ that of $\mathcal{M'}(Y)$.
Consider the PLD $\omega^c_{X/Y}$ of the composition of $\mathcal{M}$ and $\mathcal{M'}$ 
(either $\mathcal{M} \circ \mathcal{M'}$ or $\mathcal{M'} \circ \mathcal{M}$).
Denote by $\omega_{X/Y}$ the PLD of $\mathcal{M}(X)$ over $\mathcal{M}(Y)$ and 
by $\omega_{X'/Y'}$ the PLD of $\mathcal{M'}(X)$ over $\mathcal{M'}(Y)$.
The density function of $\omega^c_{X/Y}$ is given by
$$
\omega^c_{X/Y}(s)  = \int\limits_{-\infty}^\infty  \omega_{X/Y}(t) \omega_{X'/Y'}(s-t) \, \dd t.
$$
\begin{proof}
We first show that the privacy loss function of a composition is a sum of privacy loss functions.
Let $\mathcal{L}^c_{X/Y}$ denote the privacy loss function of the composition mechanism. Then,
\begin{equation} \label{eq:loss_additivity}
	\begin{aligned}
	\mathcal{L}^c_{X/Y}(t_1,t_2)	 
		=& \log \left( \frac{  f_{X,X'} (t_1,t_2) }{ f_{Y,Y'} (t_1,t_2) } \right) 
		= \log \left( \frac{ f_{X} (t_1)  f_{X'} (t_2) }{  f_{Y} (t_1)  f_{Y'} (t_2)  }\right) \\
		 = & \log \left( \frac{ f_{X} (t_1)) }{   f_{Y} (t_1)  )  }\right) 	
		+ \log \left( \frac{ f_{X'} (t_2) }{   f_{Y'} (t_2)   }\right) \\
		= & \mathcal{L}_{X/Y}(t_1) + \mathcal{L}_{X'/Y'}(t_2).
	\end{aligned}
\end{equation}
Let $S \in \mathbb{R}$ be a measurable set. By using the property \eqref{eq:loss_additivity} and by change of variables we see that
\begin{equation*}
	\begin{aligned}
	\omega^C_{X/Y}(S) &= \iint \limits_{ \{ (t_1,t_2) \in \mathbb{R}^2 \, : \, \mathcal{L}_{c}(t_1,t_2) \, \in S \} }
	f_{X,X'} (t_1,t_2) \, \dd t_1 \, \dd t_2	 \\
	 &= \iint \limits_{ \{ (t_1,t_2) \in \mathbb{R}^2 \, : \, \mathcal{L}_{X/Y}(t_1) + \mathcal{L}_{X'/Y'}(t_2) \, \in S \} }
	f_X(t_1) f_{X'} (t_2) \, \dd t_1 \, \dd t_2	 \\
	 &= \iint \limits_{ \{  s_1 + s_2 \, \in S \} \, \cap \, \{ \mathcal{L}_{X/Y}(\mathbb{R}) + \mathcal{L}_{X'/Y'}(\mathbb{R}) \}  }
	f_X\big( \mathcal{L}_{X/Y}^{-1}(s_1)  \big)  \,  \frac{\dd \mathcal{L}_{X/Y}^{-1}(s_1)}{\dd s} \, \cdot \\
	& \quad \quad \quad \quad \quad \quad f_{X'}\big( \mathcal{L}_{X'/Y'}^{-1}(s_2)  \big)  \,  \frac{\dd \mathcal{L}_{X'/Y'}^{-1}(s_2)}{\dd s} \, \dd s_1 \, \dd s_2	\\
	 &= \iint \limits_{ \{  s_1 + s_2 \, \in S \} }
	 \omega_{X/Y} (s_1) \omega_{X'/Y'} (s_2) \, \dd s_1 \, \dd s_2 \\
	 &= \int\limits_S \left( \,\, \int\limits_{-\infty}^\infty  
	  \omega_{X/Y} (s_1)  \omega_{X'/Y'} (t - y_1) \, \dd s_1 \right) \, \dd t.
	\end{aligned}
\end{equation*}
\end{proof}
\end{thm}
From Corollary~\ref{cor:delta} and Theorem~\ref{thm:convolutions} we get the following integral formula for $\delta(\veps)$.
\begin{cor}
Consider $k$ consecutive applications of a mechanism $\mathcal{M}$. Let $\veps > 0$. The composition is tightly $(\veps,\delta)$-DP for $\delta$ given by
$$
\delta(\veps) = \max_{X \sim Y} \max \{ \delta_{X/Y}(\veps), \delta_{Y/X}(\veps) \},
$$ 
where
\begin{equation*} 
	\begin{aligned}
		\delta_{X/Y}(\veps) = 
		\int\limits_\veps^\infty (1 - \ee^{\veps - s})\left( \omega_{X/Y} *^k  \omega_{X/Y}\right) (s)  \, \dd s,
	\end{aligned}
\end{equation*}
where $(\omega_{X/Y} *^k \omega_{X/Y}) (s)$ denotes the density function 
obtained by convolving $\omega_{X/Y}$ by itself $k$ times (an analogous formula holds for $\delta_{Y/X}(\veps)$).
\end{cor}

We also give the following result for the relation between $\omega_{X/Y}$ and $\omega_{Y/X}$. This result can be used to determine which the value $\max \{ \delta_{X/Y}, \delta_{Y/X} \}$.
This result can be seen as a continuous version of Lemma 2 in~\cite{sommer2019privacy}.

\begin{lem} \label{lem:dual}
Let the privacy loss functions $\mathcal{L}_{X/Y}$ and $\mathcal{L}_{Y/X}$ and the privacy loss distributions $\omega_{X/Y}$ and $\omega_{Y/X}$.
Then, it holds $\mathcal{L}_{Y/X}(\mathbb{R}) = \{ \; t \in \mathbb{R} \; : \; -t \in \mathcal{L}_{X/Y}(\mathbb{R})\}$ and
for all $y \in \mathcal{L}_{X/Y}(\mathbb{R})$:
$$
\omega_{X/Y}(s) = \ee^s \omega_{Y/X}(-s).
$$
\begin{proof}
From the definition it follows that
$$
\mathcal{L}_{X/Y}(t) = -\mathcal{L}_{Y/X}(t),
$$
and therefore also 
\begin{equation} \label{eq:L}
\mathcal{L}^{-1}_{Y/X}(s) =  \mathcal{L}^{-1}_{X/Y}(-s)
\end{equation}
for all $y \in \mathcal{L}_{X/Y}(\mathbb{R})$. Let $y \in \mathcal{L}_{X/Y}(\mathbb{R})$. Then,
\begin{equation} \label{eq:dwdy}
	\begin{aligned}
		\omega_{X/Y}(s) &= f_X\big( \mathcal{L}_{X/Y}^{-1}(s)  \big)  \,  \frac{\dd \mathcal{L}_{X/Y}^{-1}(s)}{\dd s} \\
		&= f_X\big( \mathcal{L}_{X/Y}^{-1}(s)  \big)  \, \frac{1}{  \mathcal{L}_{X/Y}'( \mathcal{L}_{X/Y}^{-1}(s) )}.		
	\end{aligned}
\end{equation}
We notice that
\begin{equation*}
	\begin{aligned}
\frac{f_X(t)}{\mathcal{L}_{X/Y}'(t)} & = \frac{f_X(t)}{ \frac{f_X'(t)}{f_X(t)} -  \frac{f_Y'(t)}{f_Y(t)}  } \\
&= \frac{f_X(t)^2 f_Y(t) }{f_X'(t)f_Y(t) - f_Y'(t)f_X(t)} \\
& = \frac{f_X(t)}{f_Y(t)}  \frac{f_X(t) f_Y(t)^2 }{f_X'(t)f_Y(t) - f_Y'(t)f_X(t)} \\
&= \ee^{\mathcal{L}_{X/Y}(t)} \frac{f_Y(t)}{\mathcal{L}_{Y/X}'(t)}
	\end{aligned}
\end{equation*}
and the claim follows using \eqref{eq:dwdy} and \eqref{eq:L}.
\end{proof}
\end{lem}
One easily verifies the following corollary of Lemma~\ref{lem:dual}.
\begin{cor}
For the convolutions it holds
$$
\Big(\omega_{X/Y} \ast^k \omega_{X/Y} \Big)(s) = 
\ee^s \, \Big( \omega_{Y/X} \ast^k \omega_{Y/X} \Big)(-s).
$$
\end{cor}

\section{Tight privacy bounds for the Gaussian mechanism via one dimensional distributions}

In this Section we show that the tight bounds of DP-SGD can be carried out by analysis of one dimensional mixture distributions.
This equivalence has also been used in~\cite[Proof of Lemma 3]{Abadi2016}.
We consider three different subsampling methods: sampling without replacement, sampling with replament and Poisson subsampling
(see~\cite{balle2018subsampling} for further details).

In the next subsection we also rigorously show that tight privacy bounds for DP-SGD can be obtained from the analysis of one dimensional distributions.

\subsection{Equivalence of the privacy bounds between the multidimensional and \\ one dimensional mechanisms}

As an example, we consider the Poisson subsampling.
In this case each member of the dataset is included in the stochastic gradient minibatch with probability $q$.
This means that each data element can appear at most once in the sample.
The basic mechanism $\mathcal{M}$ is then of the form
$$
\mathcal{M}(X) = \sum_{x \in B} f(x) + \mathcal{N}(0, \sigma^2 I_d),
$$
where $B$ is a randomly drawn subset of $\{x_1,\ldots,x_N\}$ and $\norm{f(x)}_2 \leq 1$ for all $x \in B$.

Consider the case of remove/add relation $\sim_R$ and let $X$ and $Y$ be neighbouring datasets. 
Consider first the case $q=1$, i.e., $\abs{B}=N$.  The condition of $(\veps,\delta)$-differential 
privacy states that for every measurable set $S \subset \mathbb{R}^d$ and every neighbouring $X$ and $Y$:
\begin{equation} \label{eq:cond1}
\mathbb{P}( \mathcal{M}(X) \in S ) \leq \ee^\varepsilon \mathbb{P} (\mathcal{M}(Y) \in S ) + \delta.
\end{equation}
Suppose $X = Y \cup \{x'\}$ and assume $\norm{f(x')}_2 = 1$.
and we easily see that this is then equivalent to the condition that for every measurable set $S \subset \mathbb{R}^d$:
\begin{equation} \label{eq:cond2}
\mathbb{P} \big( \mathcal{N}(f(x)',\sigma^2 I_d) \in S \big) \leq \ee^\varepsilon \mathbb{P} \big( \mathcal{N}(0,\sigma^2 I_d)  \in S \big) + \delta.
\end{equation}
Let $U \in \mathbb{R}^{d \times d}$ be a unitary matrix such that 
$$
Uf(x') = \begin{bmatrix} 1 \\ 0 \\ \vdots \\ 0 \end{bmatrix} =: e_1.
$$
This means that $U$ is of the form $U = \begin{bmatrix} f(x') & \widetilde{U} \end{bmatrix}$, 
where $\widetilde{U}$ can be taken as any $d \times (d-1)$ matrix with orthonormal columns such that $\widetilde{U}^T f(x')=0$.

Due to the unitarity of $U$, the condition \eqref{eq:cond2} is equivalent to the condition that for every measurable set $S \subset \mathbb{R}^d$:
\begin{equation} \label{eq:cond3}
	\mathbb{P} \big( U \mathcal{N}(f(x'),\sigma^2 I_d) \in U S \big) \leq \ee^\varepsilon \mathbb{P} \big( U \mathcal{N}(0,\sigma^2 I_d)  \in U S \big) + \delta.
\end{equation}	
Furthermore, due to the unitarity of $U$, $U \mathcal{N}(0,\sigma^2 I_d) \sim \mathcal{N}(0,\sigma^2 I_d)$ and we see that \eqref{eq:cond3} is equivalent to
the condition that for every measurable set $S \subset \mathbb{R}^d$:
\begin{equation} \label{eq:cond4}
\mathbb{P} \big( \mathcal{N}(e_1,\sigma^2 I_d) \in U S \big) \leq \ee^\varepsilon \mathbb{P} \big( \mathcal{N}(0,\sigma^2 I_d)  \in U S \big) + \delta, 
\end{equation}
where $US = \{ Ux \, : \, x \in S \}.$ Then, we see that the condition \eqref{eq:cond3} is equivalent to the condition that
for every measurable set $S \subset \mathbb{R}$:
\begin{equation} \label{eq:1d_cond}
	\mathbb{P} \big( \mathcal{N}(1,\sigma^2) \in S \big) \leq \ee^\varepsilon \mathbb{P} \big( \mathcal{N}(0,\sigma^2)  \in S \big) + \delta.
\end{equation}
Thus, if $X$ and $Y$ are given as above, finding the parameters $\veps$ and $\delta$ that satisfy \eqref{eq:cond1} amounts 
to finding values of $\veps$ and $\delta$ that satisfy \eqref{eq:1d_cond}.

When $q<1$, we see that $f(x')$ is in $B$ with a probability $q$.
Reasoning as above, we arrive at the the condition that for every measurable set $S \subset \mathbb{R}^d$:
\begin{equation*} 
\mathbb{P} \big( \, q \, \mathcal{N}(f(x'),\sigma^2 I_d) + (1-q) \, \mathcal{N}(0,\sigma^2 I_d) \in S \big) 
\leq \ee^\varepsilon \mathbb{P} \big( \mathcal{N}(0,\sigma^2 I_d)  \in S \big) + \delta,
\end{equation*}
where $q \mathcal{N}(f(x'),\sigma^2 I_d) + (1-q)\mathcal{N}(0,\sigma^2 I_d)$ denotes a mixture distribution.
Similarly, this leads to considering the one dimensional neighbouring distributions
$$
f_X := q \mathcal{N}(1,\sigma^2) + (1-q)\mathcal{N}(0,\sigma^2) \quad \textrm{and} \quad  f_Y := \mathcal{N}(0,\sigma^2).
$$
In order the condition \eqref{eq:cond1} holds for all $X \sim_R Y$, then it has to hold that
for every measurable set $S \subset \mathbb{R}$ both
$$
	\mathbb{P} \big( f_X \in S \big) \leq \ee^\varepsilon \mathbb{P} \big( f_Y  \in S \big) + \delta
\quad \textrm{and} \quad
	\mathbb{P} \big( f_Y \in S \big) \leq \ee^\varepsilon \mathbb{P} \big( f_X  \in S \big) + \delta.
$$

With an analogous reasoning, we see that in the case of substitution relation $\sim_S$ the worst case is obtained by considering the neighbouring distributions 
$$
q \mathcal{N}(1,\sigma^2) + (1-q)\mathcal{N}(0,\sigma^2)
$$ 
and 
$$
q \mathcal{N}(-1,\sigma^2) + (1-q)\mathcal{N}(0,\sigma^2). 
$$

Finally, we note that the case $\norm{f(x')}_2 < 1$ would lead to neighbouring distributions $f_X$ and $f_Y$ that are closer 
to each other  than in the case $\norm{f(x')}_2 = 1$. This would give tighter $(\veps,\delta)$-values, i.e., $\norm{f(x')}_2 = 1$ gives the worst case. 
This could be shown rigorously by scaling the parameter $\sigma$ and considering the analysis below. 

\subsection{Poisson subsampling}

\subsubsection{Neighbouring relation with remove/add} \label{subsec:p}

As shown above, for the analysis in case of Poisson subsampling
it is sufficient to consider the density functions (see also \cite{wang2019} and \cite{meiser2018tight})
\begin{equation} \label{eq:neigh1}
	\begin{aligned}
		f_X(t) &= q \, \tfrac{1}{\sqrt{2 \pi \sigma^2}} \ee^{ \frac{-(t-1)^2}{2 \sigma^2} }  +
		 (1-q) \tfrac{1}{\sqrt{2 \pi \sigma^2}} \ee^{ \frac{-t^2}{2 \sigma^2}}, \\
 	f_Y(t) &=  \tfrac{1}{\sqrt{2 \pi \sigma^2}} \ee^{ \frac{-t^2}{2 \sigma^2}} .
	\end{aligned}
\end{equation}
The privacy loss function $\mathcal{L}_{X/Y}(t)$ is then given by
\begin{equation*}
	\begin{aligned}
\mathcal{L}_{X/Y}(t)  = \log \frac{q \, \tfrac{1}{\sqrt{2 \pi \sigma^2}} \ee^{ \frac{-(t-1)^2}{2 \sigma^2} }  +
 (1-q) \tfrac{1}{\sqrt{2 \pi \sigma^2}} \ee^{ \frac{-t^2}{2 \sigma^2} }  }{ \tfrac{1}{\sqrt{2 \pi \sigma^2}} \ee^{ \frac{-t^2}{2 \sigma^2}} }
 = \log \left( q \, \ee^{ \frac{2t-1}{2 \sigma^2} }  + (1-q) \right).
	\end{aligned}
\end{equation*}
We see that $\mathcal{L}_{X/Y}(\mathbb{R}) = ( \log(1-q), \infty)$ and that
$\mathcal{L}_{X/Y}$ is a strictly increasing continuously differentiable
function in the whole $\mathbb{R}$. Straightforward calculation shows that
$$
\mathcal{L}_{X/Y}^{-1}(s)  = \sigma^2 \log \frac{\ee^s  - (1-q)}{ q } + \frac{1}{2}
$$
and
$$
\frac{\dd}{\dd \, s} \mathcal{L}_{X/Y}^{-1}(s) = \frac{ \sigma^2 \ee^s }{ \ee^s  - (1-q) }.
$$
The privacy loss distribution $\omega_{X/Y}$ is then given by the density function
\begin{equation*}
	\frac{\dd \, \omega_{X/Y}}{\dd \, s}(s)  =
	\begin{cases}
		f_X( \mathcal{L}_{X/Y}^{-1} (s) ) \frac{\dd}{\dd \, s} \mathcal{L}_{X/Y}^{-1}(s), &\text{ if } s > \log(1-q), \\
		0, &\text{ else. }
	\end{cases}
\end{equation*}

The privacy loss distribution $\tfrac{\dd \omega_{X/Y}}{\dd s}$ has its mass mostly on the positive real axis (equals zero for $y \leq \log(1-q)$) and so do the
the convolutions $\tfrac{\dd \omega_{X/Y}}{\dd s} \ast^k \tfrac{\dd \omega_{X/Y}}{\dd s} $.
Therefore, by Lemma~\ref{lem:dual} and its corollary, we see that
$\tfrac{\dd \omega_{Y/X}}{\dd s}$ has its mass mostly on the negative real axis (equals zero for $y \geq \abs{\log(1-q)}$).
Thus the representation \eqref{eq:pldrepr2} supports the numerical observation that generally
$\delta = \delta_{X/Y}$.


\subsection{Sampling without replacement and $\sim_S$-neighouring relation}  \label{subsec:bounded}


Denote by $m$ the batch size (fixed) and $q=m/N$.
In case of sampling without replacement and $(\varepsilon, \delta, \sim_S)$-DP, the differing element is in the minibatch with a probability $q$,
and without loss of generality, we may again consider the density functions
\begin{equation} \label{eq:neigh2}
	\begin{aligned}
		f_X(t) &= q \, \tfrac{1}{\sqrt{2 \pi \sigma^2}} \ee^{ \frac{-(t-1)^2}{2 \sigma^2} }  +
		 (1-q) \tfrac{1}{\sqrt{2 \pi \sigma^2}} \ee^{ \frac{-t^2}{2 \sigma^2}}, \\
 	f_Y(t) &=  q \, \tfrac{1}{\sqrt{2 \pi \sigma^2}} \ee^{ \frac{-(t+1)^2}{2 \sigma^2} }  +
		 (1-q) \tfrac{1}{\sqrt{2 \pi \sigma^2}} \ee^{ \frac{-t^2}{2 \sigma^2}}.
	\end{aligned}
\end{equation}
The privacy loss function is then given by
\begin{equation*}
	\begin{aligned}
\mathcal{L}_{X/Y}(t) = \log \left( \frac{q \, \tfrac{1}{\sqrt{2 \pi \sigma^2}} \ee^{ \frac{-(t-1)^2}{2 \sigma^2} }  + 
(1-q) \tfrac{1}{\sqrt{2 \pi \sigma^2}} \ee^{ \frac{-t^2}{2 \sigma^2} }  }
{q \, \tfrac{1}{\sqrt{2 \pi \sigma^2}} \ee^{ \frac{-(t+1)^2}{2 \sigma^2} }  + (1-q) \tfrac{1}{\sqrt{2 \pi \sigma^2}} \ee^{ \frac{-t^2}{2 \sigma^2} } } \right) 
 = \log \left( \frac{ q \, \ee^{ \frac{2t-1}{2 \sigma^2} }  + (1-q) }{q \, \ee^{ \frac{-2t-1}{2 \sigma^2} }  + (1-q) }    \right).  
	\end{aligned}
\end{equation*}
Now $\mathcal{L}_{X/Y}(\mathbb{R}) = \mathbb{R}$ and $\mathcal{L}_{X/Y}$ is again a strictly increasing continuously differentiable
function in the whole $\mathbb{R}$. Denote 
$$
x =\ee^{\frac{t}{\sigma^2}} \quad \textrm{and} \quad c = q \ee^{- \frac{1}{2 \sigma^2}}.
$$ 
Then, solving $\mathcal{L}_{X/Y}(t) = s$ leads to the equation
\begin{equation*}
	\begin{aligned}
		& \frac{ cx + (1-q) }{cx^{-1} + (1-q)} = \ee^s \\
		\iff \quad & cx^2 + (1-q)(1-\ee^s)x - c \ee^s  = 0 \\
		\overset{x>0}{\iff} \quad & x = \frac{ -(1-q)(1-\ee^s) + \sqrt{ (1-q)^2(1-\ee^s)^2 + 4c^2 \ee^s } }{ 2c  }.
	\end{aligned}
\end{equation*}
We find that
\begin{equation*}
	\mathcal{L}_{X/Y}^{-1}(s) = \sigma^2 \log \Bigg(   \frac{ -(1-q)(1-\ee^s)  + \sqrt{ (1-q)^2(1-\ee^s)^2 + 4c^2 \ee^s } }{ 2c  }  \Bigg)
\end{equation*}
and
$$
\frac{\dd}{\dd \, s} \mathcal{L}_{X/Y}^{-1}(s) =  
\sigma^2  \frac{ \frac{4c^2 \ee^s - 2(1-q)^2 \ee^s (1-\ee^s)}{ 2 \sqrt{ 4c^2\ee^s + (1-q)^2(1-\ee^s)^2 } } + (1-q) \ee^s }
{  \sqrt{ 4c^2\ee^s + (1-q)^2(1-\ee^s)^2 } - (1-q)(1-\ee^s) }.
$$
In case of odd loss functions ($f_Y(-t) = f_X(t)$ and $\mathcal{L}_{X/Y}(-t) = - \mathcal{L}_{X/Y}(t)$)
we have the following:
$$
\frac{\dd \omega_{X/Y}}{\dd s}(s) = \frac{\dd \omega_{Y/X}}{\dd s}(s).
$$
This follows from using the oddity of $\mathcal{L}_{X/Y}$ and Lemma~\ref{lem:dual}.
Therefore, if $f_Y(-t) = f_X(t)$ and $\mathcal{L}_{X/Y}(-t) = - \mathcal{L}_{X/Y}(t)$,
it holds $\delta = \delta_{Y/X} = \delta_{X/Y}$ by the representation \eqref{eq:pldrepr2}.

We remark that in $(\varepsilon, \delta, \sim_S)$-DP, the Poisson subsampling with the sampling parameter $\gamma$ (i.e., each sample is in the batch with a probability $\gamma$) 
is equivalent to the case of the sampling with replacement with $q=\gamma$, as in both cases the differing element is included in the minibatch with probability $\gamma$.


\subsection{Sampling with replacement and $\sim_S$-neighouring relation}  \label{subsec:bounded2}


Consider next the sampling with replacement and the $\sim_S$-neighouring relation. Then the number of times the differing sample $x'$ is in the batch is binomially distributed,
i.e., the probability for being in the batch $\ell$ times is $\Big( \frac{1}{n} \Big)^\ell \Big( \frac{1}{n} \Big)^{m-\ell} {m\choose\ell}$,
where $m$ denotes the batch size and $n$ the total number of data samples.

Then, without loss of generality, we may consider the density functions ($m$ denotes the batch size)
\begin{equation} \label{eq:wr}
	\begin{aligned}
		f_X(t) &=  \tfrac{1}{\sqrt{2 \pi \sigma^2}}  \sum\limits_{\ell=0}^m q^\ell (1-q)^{m-\ell} {m\choose\ell} \,\ee^{ \frac{-(t-\ell)^2}{2 \sigma^2} }, \\
 	f_Y(t) &=   \tfrac{1}{\sqrt{2 \pi \sigma^2}}  \sum\limits_{\ell=0}^m q^\ell (1-q)^{m-\ell} {m\choose\ell} \,\ee^{ \frac{-(t+\ell)^2}{2 \sigma^2} },
	\end{aligned}
\end{equation}
where $q=1/n$.
The privacy loss function is then given by
\begin{equation*}
	\begin{aligned}
\mathcal{L}_{X/Y}(t) = \log \left( \frac{ \sum\limits_{\ell=0}^m q^\ell (1-q)^{m-\ell}  {m\choose\ell} \,\ee^{ \frac{-(t-\ell)^2}{2 \sigma^2} } }{ \sum\limits_{\ell=0}^m q^\ell (1-q)^{m-\ell} {m\choose\ell}\,\ee^{ \frac{-(t+\ell)^2}{2 \sigma^2} } } \right) 
 =  \log \left( \frac{ \sum\limits_{\ell=0}^m  c_\ell x^\ell }{ \sum\limits_{\ell=0}^m  c_\ell x^{-\ell}} \right), 
	\end{aligned}
\end{equation*}
where 
\begin{equation} \label{eq:c_x}
	c_\ell = q^\ell (1-q)^{m-\ell}  {m\choose\ell} \,\ee^{  \frac{ - \ell^2}{2 \sigma^2} } \quad \textrm{and} \quad x = \ee^{  \frac{ t }{ \sigma^2} }.
\end{equation}
Since $c_\ell>0$ for all $\ell = 1, \ldots, m$, clearly $\sum_{\ell=0}^m  c_\ell x^\ell$ is strictly increasing as a function of $t$ and
$\sum_{\ell=0}^m  c_\ell x^{-\ell}$ is strictly decreasing. Moreover, we see that 
$$
\tfrac{ \sum_{\ell=0}^m  c_\ell x^\ell }{ \sum_{\ell=0}^m  c_\ell x^{-\ell}} \rightarrow 0 \quad \textrm{as} \,\, t \rightarrow -\infty
\quad \quad \textrm{and} \quad \quad
\tfrac{ \sum_{\ell=0}^m  c_\ell x^\ell }{ \sum_{\ell=0}^m  c_\ell x^{-\ell}}  \rightarrow \infty \quad \textrm{as} \,\,t \rightarrow \infty.
$$ 
Thus, $\mathcal{L}_{X/Y}(\mathbb{R}) = \mathbb{R}$ and
$\mathcal{L}_{X/Y}(t)$ is a strictly increasing continuously differentiable function in the whole $\mathbb{R}$.
To find $\mathcal{L}_{X/Y}^{-1}(s)$ one needs to solve $\mathcal{L}_{X/Y}(t) = s$, i.e., one needs to find the single real root of a polynomial of order $2m$.

To find $\mathcal{L}_{X/Y}^{-1}(s)$, i.e. to solve $\mathcal{L}_{X/Y}(t) = s$ for a given $y$, one may use e.g. Newton's method.

\section{Error estimates} \label{sec:err_est_a}

For the error analysis we consider the Poisson subsampling with $(\varepsilon, \delta, \sim_R)$-DP, i.e., we consider
the PLD density function (Sec.~\ref{subsec:p})
\begin{equation} \label{eq:e_case}
\omega(s) = \begin{cases}
f(g(s))g'(s), &\text{ if }  s > \log(1-q), \\
0, &\text{ otherwise},
\end{cases}
\end{equation}
where
\begin{equation*} 
f(t) = \frac{1}{\sqrt{2 \pi \sigma^2}} \, [ q \ee^{ \frac{-(t-1)^2}{2 \sigma^2}} + (1-q) \ee^{-\frac{t^2}{2 \sigma^2}} ],
\end{equation*}
\begin{equation*} 
g(s) = \sigma^2 \log \left( \frac{\ee^s - (1-q)}{q} \right) + \frac{1}{2}.
\end{equation*}

\begin{thm} \label{thm:total_error}
Let the vector $C^k$ be defined as in Sec.~\ref{subsec:desc}.
Total error  of the approximation (determined by the truncation parameter $L$ and the discretisation parameter $n$) can be bounded by three terms as follows:
\begin{equation*} 
	\begin{aligned}
   \abs{\int\limits_\veps^\infty (1 - \ee^{\veps - s})(\omega *^k \omega ) (s)  \, \dd s - 
 \Delta x \sum\limits_{\ell=0}^{n-1}  \big(1 - \ee^{\veps - ( \ell \Delta x)} \big) C_\ell^k} 
\leq  I_1(L) + I_2(L) + I_3(L,n),
	\end{aligned}
\end{equation*}
where
\begin{equation*} 
	\begin{aligned}
 I_1(L) &  =  \int\limits_L^\infty (\omega *^k \omega ) (s)  \, \dd s, \\
 I_2(L) & =  \abs{\int\limits_\veps^L (\omega *^k \omega ) (s) - (\widetilde{\omega} \circledast^k \widetilde{\omega} ) (s)  \, \dd s }, \\
 I_3(L,n) & = \abs{\int\limits_\veps^L (1 - \ee^{\veps - s})(\widetilde{\omega} \circledast^k \widetilde{\omega} ) (s)  \, \dd s
		- \Delta x \sum\limits_{\ell=0}^{n-1}  \big(1 - \ee^{\veps - ( \ell \Delta x)} \big) C_\ell^k}.
	\end{aligned}
\end{equation*}
\begin{proof}
By adding and subtracting terms and using the triangle inequality, we get
\begin{equation} \label{eq:total_error}
	\begin{aligned}
&    \abs{\int\limits_\veps^\infty (1 - \ee^{\veps - s})(\omega *^k \omega ) (s)  \, \dd s - 
 \Delta x \sum\limits_{\ell=0}^{n-1}  \big(1 - \ee^{\veps - ( \ell \Delta x)} \big) C_\ell^k}  \\
& \quad \leq 	 \abs{ \int\limits_\veps^\infty (1 - \ee^{\veps - s})(\omega *^k \omega ) (s)  \, \dd s -
		\int\limits_\veps^L (1 - \ee^{\veps - s})(\omega *^k \omega ) (s)  \, \dd s } \\ 
&	\quad + \abs{\int\limits_\veps^L (1 - \ee^{\veps - s})(\omega *^k \omega ) (s)  \, \dd s -
		\int\limits_\veps^L (1 - \ee^{\veps - s})(\widetilde{\omega} \circledast^k \widetilde{\omega} ) (s)  \, \dd s } \\ 
&  \quad	+ \abs{\int\limits_\veps^L (1 - \ee^{\veps - s})(\widetilde{\omega} \circledast^k \widetilde{\omega} ) (s)  \, \dd s
		- \Delta x \sum\limits_{\ell=0}^{n-1}  \big(1 - \ee^{\veps - ( \ell \Delta x)} \big) C_\ell^k} \\
 &	 \quad \leq \int\limits_L^\infty (\omega *^k \omega ) (s)  \, \dd s 
	+ \abs{\int\limits_\veps^L (\omega *^k \omega ) (s) - (\widetilde{\omega} \circledast^k \widetilde{\omega} ) (s)  \, \dd s } \\
& 	\quad + \abs{\int\limits_\veps^L (1 - \ee^{\veps - s})(\widetilde{\omega} \circledast^k \widetilde{\omega} ) (s)  \, \dd s
		- \Delta x \sum\limits_{\ell=0}^{n-1}  \big(1 - \ee^{\veps - ( \ell \Delta x)} \big) C_\ell^k}.
	\end{aligned}
\end{equation}
\end{proof}
\end{thm}

We consider next separately each of the three terms on the right hand side of \eqref{eq:total_error}.

\subsection{Tail bounds for the convolved PLDs}

The first term on the right hand side of \eqref{eq:total_error} is bounded by the tail of the convolved PLDs:
\begin{equation} \label{eq:tail}
\abs{ \int\limits_\veps^\infty (1 - \ee^{\veps - s})(\omega *^k \omega ) (s)  \, \dd s -
		\int\limits_\veps^L (1 - \ee^{\veps - s})(\omega *^k \omega ) (s)  \, \dd s } \leq 
		\int\limits_L^\infty (\omega *^k \omega ) (s)  \, \dd s.
\end{equation}
In this Section we show how to use existing R\'enyi differential privacy (RDP) results to bound the tail \eqref{eq:tail}.

The Chernoff bound (see e.g. \cite{wainwright2019}) states that for any random variable $X$ and for all $\lambda > 0$ it holds
\begin{equation} \label{eq:Markov}
\mathbb{P}[ X \geq t] = \mathbb{P}[ \ee^{\lambda X} \geq \ee^{\lambda t} ] \leq \frac{ \mathbb{E}[ \ee^{\lambda X} ] }{\ee^{\lambda t}}.
\end{equation}
From the RDP bounds given in~\cite{mironov2019} we obtain the following bound for the moment generating function $\mathbb{E} [ \ee^{\lambda \omega}  ]$.
\begin{lem} \label{lem:mgfct}
Suppose $q\leq \frac{1}{5}$ and $\sigma \geq 4$. Suppose $\lambda$ satisfies
\begin{equation*}
	\begin{aligned}
		1 < & \lambda \leq \frac{1}{2}\sigma^2 c - 2 \log \sigma, \\
		 & \lambda \leq \frac{\frac{1}{2}\sigma^2 c  - \log \, 5 - 2 \log \, \sigma}{c + \log(q \lambda) + 1/(2\sigma^2)},
	\end{aligned}
\end{equation*}
where $c = \log \left( 1 + \frac{1}{q ( \lambda - 1)}   \right)$. Then,
$$
\mathbb{E} [ \ee^{\lambda \omega}  ] \leq 1 + \frac{2q^2(\lambda+1)\lambda}{\sigma^2}.
$$
\begin{proof}
Making change of variables $y = \mathcal{L}(t)$ (recall: $\mathcal{L}(\mathbb{R}) = (\log(1-q),\infty)$ and
$\mathcal{L}(t)$ is a strictly increasing differentiable function),
we see a connection to the R\'enyi differential privacy:
\begin{equation*}
	\begin{aligned}
	\mathbb{E} [ \ee^{\lambda \omega}  ] & = \int\limits_{\log(1-q)}^\infty \ee^{\lambda s} \omega (s) \, \dd s\\
	& = \int\limits_{-\infty}^\infty \ee^{\lambda \mathcal{L}(t) } f_X(t) \, \dd t \\
	& = \int\limits_{-\infty}^\infty \left( \frac{f_X(t)}{f_Y(t)}  \right)^\lambda f_X(t) \, \dd t \\
	& = \int\limits_{-\infty}^\infty \left( \frac{f_X(t)}{f_Y(t)}  \right)^{\lambda + 1}  f_Y(t) \, \dd t.
	\end{aligned}
\end{equation*}
Here $f_X(t) = q \mu_1(t) + (1-q) \mu_0(t)$, where $\mu_0(t) = \frac{1}{\sqrt{2 \pi \sigma^2}}  \ee^{ \frac{-t^2}{2 \sigma^2}} $ and
$\mu_1(t) = \frac{1}{\sqrt{2 \pi \sigma^2}}  \ee^{ \frac{-(t-1)^2}{2 \sigma^2}}$, and 
$f_Y(t) = \mu_0(t)$.  Therefore
\begin{equation} \label{eq:expr}
	\begin{aligned}
\mathbb{E} [ \ee^{\lambda \omega}  ] = 
\int\limits_{-\infty}^\infty \left( \frac{f_X(t)}{f_Y(t)}  \right)^{\lambda + 1}  f_Y(t) \, \dd t = 
\int\limits_{-\infty}^\infty \left( (1-q) + q \frac{\mu_1(t)}{\mu_0(t)} \right)^{\lambda + 1} \mu_0(t) \, \dd t.
	\end{aligned}
\end{equation}
From the proof of~\cite[Thm.\;11]{mironov2019} we get a bound for \eqref{eq:expr} which shows the claim.
\end{proof}
\end{lem}

\begin{thm} \label{thm:S_k_bound}
Let the assumptions on $\sigma$ and $q$  of Lemma~\ref{lem:mgfct} hold. Assume $\omega_i$, $i=1,\ldots,k$ are independent PLDs of the form \eqref{eq:e_case}
determined by $\sigma$ and $q$. Denote $S_k := \sum_{i=1}^k \omega_i$. Then, it holds
$$
\mathbb{P}( S_k \geq L) \leq  \left(  1 + \frac{2q^2(\lambda+1)\lambda}{\sigma^2}  \right)^k \ee^{-L \lambda}
$$
for all $\lambda$ that satisfy the assumptions of Lemma~\ref{lem:mgfct}.
\begin{proof}
Since $\omega_i$'s are independent, we have by Lemma~\ref{lem:mgfct},
$$
\mathbb{E} [ \ee^{\lambda S_k}  ]  = \prod_{i=1}^k \mathbb{E} [ \ee^{\lambda \omega_i}  ] \leq  \left(  1 + \frac{2q^2(\lambda+1)\lambda}{\sigma^2}  \right)^k.
$$
Using the Chernoff bound, we find that
$$
\mathbb{P}( S_k \geq L) \leq  \left(  1 + \frac{2q^2(\lambda+1)\lambda}{\sigma^2}  \right)^k \ee^{-L \lambda}.
$$
For all $\lambda$ that satisfy the assumptions of Lemma~\ref{lem:mgfct}.
\end{proof}
\end{thm}
The parameter $\lambda$ in Theorem~\ref{thm:S_k_bound} can be chosen freely as long as it satisfies the conditions of Lemma~\ref{lem:mgfct}.
The $\lambda$ that minimises the function $\lambda^2 \ee^{-L\lambda}$ is given by $\lambda = \tfrac{L}{2}$. This choice leads to the following bound.
\begin{cor}
Let $L$ be chosen such that $\lambda = L/2$ satisfies the assumptions of Lemma~\ref{lem:mgfct}. Then, we have the following bound:
$$
\mathbb{P}( S_k \geq L) \leq  \left(  1 + \frac{2q^2(\tfrac{L}{2}+1)\tfrac{L}{2}}{\sigma^2}  \right)^k \ee^{-\frac{L^2}{2}}.
$$
\end{cor}

Notice that 
$$
\mathbb{P}( S_k \geq L) =  \int\limits_L^\infty  ( \omega \ast^k \omega)(s) \, \dd s.
$$

\begin{figure} [h!]
	\begin{center}
  \includegraphics[width=0.5\linewidth]{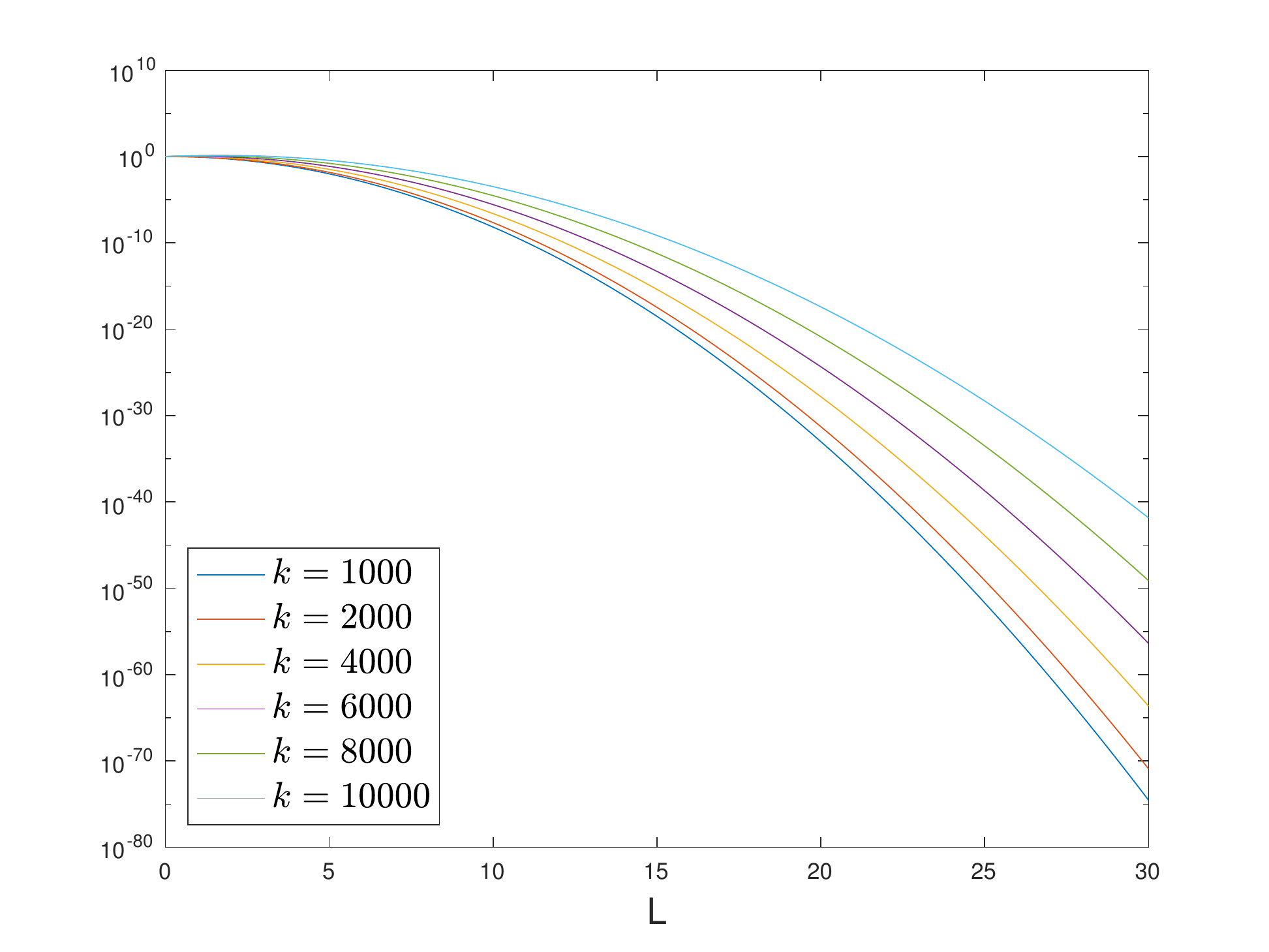}
	\caption{Convergence of the bound given by Corollary 4. }
\label{fig:conv_bound}
	\end{center}
\end{figure}

\textbf{Example.} Set $q=0.01$, $\sigma=4.0$. We numerically observe that the conditions of Lemma 2 hold up to $\lambda \approx 14.3$.
Thus, Corollary 4 holds up to $L\approx 28.6$. Figure~\ref{fig:conv_bound} shows the convergence of the bound with respect to $L$.

\subsection{Errors arising from truncation of the convolution integrals and periodisation} 


We next bound the second term on the right hand side of \eqref{eq:total_error}, i.e. the term
$$
\abs{\int\limits_\veps^L (1 - \ee^{\veps - s})(\omega *^k \omega ) (s)  \, \dd s -
		\int\limits_\veps^L (1 - \ee^{\veps - s})(\widetilde{\omega} \circledast^k \widetilde{\omega} ) (s)  \, \dd s }.
$$
We easily see that this can be bounded as
\begin{equation} \label{eq:period}
	\begin{aligned}
& \abs{\int\limits_\veps^L (1 - \ee^{\veps - s})(\omega *^k \omega ) (s)  \, \dd s - 
		\int\limits_\veps^L (1 - \ee^{\veps - s})(\widetilde{\omega} \circledast^k \widetilde{\omega} ) (s)  \, \dd s } \\
\leq & \int\limits_\veps^L \abs{(\omega *^k \omega - \widetilde{\omega} \circledast^k \widetilde{\omega})(x)} \, \dd x  \\
\leq & \int\limits_\veps^L \abs{(\omega *^k \omega - \omega \circledast^k \omega)(x)} \, \dd x  + \int\limits_\veps^L\abs{(\omega \circledast^k \omega - \widetilde{\omega} \circledast^k \widetilde{\omega})(x)} \, \dd x.
	\end{aligned}
\end{equation}

\subsubsection{Truncation of the convolution integrals }

We first bound $\omega *^k \omega - \omega \circledast^k \omega$. We have the following pointwise bound.

\begin{lem} \label{lem:trunc}
Let $\sigma>0$ and $0<q<\frac{1}{2}$. 
Let $\omega$ be defined as above, and let $L\geq 1$.
Then, for all $x \in \mathbb{R}$,
$$
\abs{(\omega *^k \omega - \omega \circledast^k \omega)(x)} \leq k \sigma \ee^{ - \frac{ - ( \sigma^2 L + C )^2 }{2 \sigma^2}},
$$
where $C = \sigma^2 \log( \frac{1}{2q}) - \frac{1}{2}$
\begin{proof}
By adding and subtracting, we may write
\begin{equation} \label{eq:s0}
\omega *^k \omega - \omega \circledast^k \omega = \omega \circledast ( \omega *^{k-1} \omega - \omega \circledast^{k-1} \omega ) 
+ \omega \circledast ( \omega \circledast^{k-1} \omega ) - \omega * ( \omega \circledast^{k-1} \omega ),	
\end{equation}
where
\begin{equation*}
\begin{aligned}
&	\omega \circledast ( \omega \circledast^{k-1} \omega ) - \omega * ( \omega \circledast^{k-1} \omega ) (x) \\
& \, = \int\limits_{-L}^L \omega(t) (\omega \circledast^{k-1} \omega)(x-t) \, \dd t
	- \int\limits_{-\infty}^\infty  \omega(t) (\omega \circledast^{k-1} \omega)(x-t) \, \dd t \\
	& \,	 =  - \int\limits_L^\infty  \omega(t) (\omega \circledast^{k-1} \omega)(x-t) \, \dd t,
\end{aligned}
\end{equation*}
since $\omega(s) = 0$ for all $s < \log (1-q)$ and $-L < \log (1-q)$.
Using Lemma \ref{lem:boundy2} of Appendix, we see that for all $x$,
\begin{equation} \label{eq:s1}
\begin{aligned}
	\abs{ \big( \omega \circledast ( \omega \circledast^{k-1} \omega ) - \omega * ( \omega \circledast^{k-1} \omega ) \big) (x)  } & \leq \max\limits_{s \geq L} \omega(s) 
	\int\limits_L^\infty  (\omega \circledast^{k-1} \omega)(x-t) \, \dd t \\
	& \leq \max\limits_{s \geq L} \omega(s) \\
	& \leq \sigma  \ee^{ - \frac{ - ( \sigma^2 L + C )^2 }{2 \sigma^2} }.
\end{aligned}	
\end{equation}
Using again Lemma \ref{lem:boundy2}, we see that for all $x$,
\begin{equation} \label{eq:s2}
\begin{aligned}
\abs{ ( \omega * \omega - \omega \circledast \omega )(x) } & = \int\limits_{-\infty}^\infty \omega(t) \omega(x-t) \, \dd t - \int\limits_{-L}^L \omega(t) \omega(x-t) \, \dd t \\
& = \int\limits_L^\infty \omega(t) \omega(x-t) \, \dd t \\
& \leq \max\limits_{s \geq L} \omega(s) \int\limits_L^\infty \omega(x-t) \, \dd t \\
& \leq  \sigma	\ee^{ - \frac{ - ( \sigma^2 L + C )^2 }{2 \sigma^2} }.
\end{aligned}
\end{equation}
The claim follows from the recursion \eqref{eq:s0} and the bounds \eqref{eq:s1} and \eqref{eq:s2}.
\end{proof}
\end{lem}

\subsubsection{Error arising from the periodisation} \label{subsec:third_approx}

We next bound the second term on the right hand side of \eqref{eq:period}.
The bound is expressed in terms of the the log of the moment 
generating function of the privacy loss function $\mathcal{L} = \mathcal{L}_{X/Y}$ (see also ~\cite{Abadi2016}) which is defined
for all $\lambda>0$ as
$$
\alpha(\lambda) := \log \mathop{\mathbb{E}}_{t \sim f_X(t)} [\ee^{\lambda \mathcal{L}(t)}].
$$
As shown in equation \eqref{eq:pld_lmf} of the main text, $\alpha(\lambda)$ is related to the moment generating function 
of the privacy loss distribution as
\begin{equation} \label{eq:conn}
	\begin{aligned}
		\mathbb{E} [\ee^{\lambda \omega }] = \ee^{\alpha(\lambda)}.
	\end{aligned}
\end{equation}
Thus, using the Chernoff bound and \eqref{eq:conn}, tail bounds involving $\omega$ can bounded in terms of $\alpha(\lambda)$.
Bounds for $\alpha(\lambda)$ in the case of Poisson subsampling with $\sim_R$ neighbouring relation are given in~\cite{Abadi2016} and~\cite{mironov2019}.

\begin{lem} \label{lem:period}
Let $\omega$ be defined as above. Then, 
$$
\int\limits_\veps^L\abs{(\omega \circledast^k \omega - \widetilde{\omega} \circledast^k \widetilde{\omega})(x)} \, \dd x  \leq
 \ee^{\alpha(L/2)} \ee^{- \frac{L^2}{2}} + 2 \sum\limits_{n=1}^\infty \ee^{k \alpha( nL)} \ee^{-2 (nL)^2}.
$$
\begin{proof}
	
We see that
\begin{equation} \label{eq:1x}
\begin{aligned}
	 & (\widetilde{\omega} \circledast^k \widetilde{\omega}  - \omega \circledast^k \omega)(x)  \\
& \quad \quad 	=    \int\limits_{-L}^L \widetilde{\omega}(t_1) \ldots \int\limits_{-L}^L \widetilde{\omega}(t_{k-1}) \, \widetilde{\omega}(x - \sum\nolimits_{i=1}^{k-1} t_i  ) \, \dd t_1 \ldots \dd t_{k-1}  \\
	 & \quad \quad \quad \quad  - \int\limits_{-L}^L \omega(t_1) \ldots \int\limits_{-L}^L  \omega(t_{k-1}) \, \omega(x - \sum\nolimits_{i=1}^{k-1} t_i  ) \, \dd t_1 \ldots \dd t_{k-1} \\
	& \quad \quad  =  \int\limits_{-L}^L \omega(t_1) \ldots \int\limits_{-L}^L \omega(t_{k-1}) \, \widetilde{\omega}(x - \sum\nolimits_{i=1}^{k-1} t_i  ) \, \dd t_1 \ldots \dd t_{k-1} \\ 
	 	 & \quad \quad \quad \quad  - \int\limits_{-L}^L \omega(t_1) \ldots \int\limits_{-L}^L  \omega(t_{k-1}) \, \omega(x - \sum\nolimits_{i=1}^{k-1} t_i  ) \, \dd t_1 \ldots \dd t_{k-1}  \\
	& \quad \quad  = \int\limits_{-L}^L \omega(t_1) \ldots \int\limits_{-L}^L  \omega(t_{k-1}) \Big( \widetilde{\omega}(x - \sum\nolimits_{i=1}^{k-1} t_i  ) -  \omega(x - \sum\nolimits_{i=1}^{k-1} t_i  ) \Big) \, \dd t_1 \ldots \dd t_{k-1},  
\end{aligned}
\end{equation}
since $\omega = \widetilde{\omega}$ on the interval $[-L,L]$. 

Recall that $\widetilde{\omega}$ is the $2L$-periodic function for which $\widetilde{\omega}(t) = \omega(t)$ for all $t \in [-L,L]$. Therefore
\begin{equation} \label{eq:2x}
	\widetilde{\omega}(t) - \omega(t) = \sum\limits_{n \in \mathbb{Z}\setminus \{0\}} \widehat{\omega}_n(t) - r(t),
\end{equation}	
where
$$
\widehat{\omega}_n(t) = \begin{cases}
	\omega(t - 2nL), &\text{ if } t \in [(2n-1)L,(2n+1)L]\\
0, &\text{ else,}	
\end{cases}
$$
and
$$
r(t) = \begin{cases}
	\omega(t), &\text{ if } t \geq L\\
0, &\text{ else.}	
\end{cases}
$$
Thus, from \eqref{eq:1x} and \eqref{eq:2x} it follows that
\begin{equation} \label{eq:C12} 
	 (\widetilde{\omega} \circledast^k \widetilde{\omega}  - \omega \circledast^k \omega)(x) = C_1(x) + C_2(x),
\end{equation}
where
$$
C_1(x) =\int\limits_{-L}^L \omega(t_1) \ldots \int\limits_{-L}^L  \omega(t_{k-1}) \sum\limits_{n \in \mathbb{Z}\setminus \{0\}} \widehat{\omega}_n(x - \sum\nolimits_{i=1}^{k-1} t_i ) \, \dd t_1 \ldots \dd t_{k-1} 
$$
and
$$
C_2(x) =\int\limits_{-L}^L \omega(t_1) \ldots \int\limits_{-L}^L  \omega(t_{k-1})  r(x - \sum\nolimits_{i=1}^{k-1} t_i ) \, \dd t_1 \ldots \dd t_{k-1} .
$$

We see that $\abs{C_1(x) }$ can be bounded as
\begin{equation*} 
\begin{aligned}
	\abs{C_1(x) }& = \int\limits_{-L}^L \omega(t_1) \ldots \int\limits_{-L}^L  \omega(t_{k-1}) \sum\limits_{n \in \mathbb{Z}\setminus \{0\}} \widehat{\omega}_n(x - \sum\nolimits_{i=1}^{k-1} t_i ) \, \dd t_1 \ldots \dd t_{k-1}   \\
	& = \sum\limits_{n \in \mathbb{Z}\setminus \{0\}}  \int\limits_{-L}^L \omega(t_1) \ldots \int\limits_{-L}^L  \omega(t_{k-1}) \, \widehat{\omega}_n(x - \sum\nolimits_{i=1}^{k-1} t_i ) \, \dd t_1 \ldots \dd t_{k-1}   \\
	& \leq \sum\limits_{n \in \mathbb{Z}\setminus \{0\}}  \int\limits_{-L}^L \omega(t_1) \ldots \int\limits_{-L}^L  \omega(t_{k-1}) \, \omega(x - 2nL  - \sum\nolimits_{i=1}^{k-1} t_i ) \, \dd t_1 \ldots \dd t_{k-1}   \\
	& \leq \sum\limits_{n \in \mathbb{Z}\setminus \{0\}}  \int\limits_{-\infty}^\infty \omega(t_1) \ldots \int\limits_{-\infty}^\infty  \omega(t_{k-1}) \, \omega(x - 2nL  - \sum\nolimits_{i=1}^{k-1} t_i ) \, \dd t_1 \ldots \dd t_{k-1}   \\
	& = \sum\limits_{n \in \mathbb{Z}\setminus \{0\}} (\omega \ast^k \omega)(x - 2nL).
\end{aligned}
\end{equation*}
Next, consider the expression
\begin{equation} \label{eq:2nL}
 \int\limits_\veps^L \sum\limits_{n \in \mathbb{Z}\setminus \{0\}} (\omega \ast^k \omega)(x - 2nL)  \, \dd x
 = \sum\limits_{n=1}^\infty \int\limits_\veps^L (\omega \ast^k \omega)(x - 2nL)  \, \dd x 
 + \sum\limits_{n=1}^\infty \int\limits_\veps^L (\omega \ast^k \omega)(x + 2nL)  \, \dd x.	
\end{equation}
Clearly, for the second term on the right hand side of \eqref{eq:2nL},
\begin{equation} \label{eq:f1}
	\begin{aligned}
\sum\limits_{n=1}^\infty \int\limits_\veps^L (\omega \ast^k \omega)(x + 2nL)  \, \dd x & = \sum\limits_{n=1}^\infty \int\limits_{\veps+2nL}^{L+2nL} (\omega \ast^k \omega)(x)  \, \dd x \\
& \leq \sum\limits_{n=1}^\infty \int\limits_{\veps+2nL}^\infty (\omega \ast^k \omega)(x)  \, \dd x \\
& \leq \sum\limits_{n=1}^\infty \int\limits_{2nL}^\infty (\omega \ast^k \omega)(x)  \, \dd x \\
& \leq \sum\limits_{n=1}^\infty \ee^{k \alpha( nL)} \ee^{-2 (nL)^2}, \\
	\end{aligned}
\end{equation}
where on the last step we use the Chernoff bound for each term with $\lambda = nL$.

In order to bound the second term on the right hand side of \eqref{eq:2nL} we consider the following.
From the Chernoff bound we get
\begin{equation} \label{eq:3a}
	\mathbb{P}( \omega \leq -L) = \mathbb{P}( - \omega \geq L) \leq \frac{\mathbb{E} [\ee^{- \lambda \omega}]}{\ee^{\lambda L}}
\end{equation}
for all $\lambda > 0$. 

Let us use again the notation of the proof of Lemma~\ref{lem:mgfct}, i.e.,
 denote $f_X(t) = q \mu_1(t) + (1-q) \mu_0(t)$, where $\mu_0(t) = \tfrac{1}{\sqrt{2 \pi \sigma^2}}  \ee^{ \tfrac{-t^2}{2 \sigma^2}} $ and
$\mu_1(t) = \tfrac{1}{\sqrt{2 \pi \sigma^2}}  \ee^{ \tfrac{-(t-1)^2}{2 \sigma^2}}$, and 
$f_Y(t) = \mu_0(t)$.  By change of variables $s = \mathcal{L}_{X/Y}(t)$, we see that
\begin{equation} \label{eq:1a}
	\begin{aligned}
\mathbb{E} [ \ee^{ - \lambda \omega}  ] = \int\limits_{-\infty}^\infty \ee^{- \lambda \log \frac{f_X(t)}{f_Y(t)}} f_X(t) \, \dd t = 
\int\limits_{-\infty}^\infty \left( \frac{f_Y(t)}{f_X(t)}  \right)^\lambda  f_X(t) \, \dd t.
	\end{aligned}
\end{equation}
From~\cite[Corollary\;7]{mironov2019} it follows that for all $\lambda \geq 1$,
\begin{equation} \label{eq:1b}
\int\limits_{-\infty}^\infty \left( \frac{f_Y(t)}{f_X(t)}  \right)^\lambda  f_X(t) \, \dd t \leq 
\int\limits_{-\infty}^\infty \left( \frac{f_X(t)}{f_Y(t)}  \right)^\lambda  f_Y(t) \, \dd t = 
\int\limits_{-\infty}^\infty \left( \frac{f_X(t)}{f_Y(t)}  \right)^{\lambda-1}  f_X(t) \, \dd t = \mathbb{E} [ \ee^{ (\lambda-1) \omega}  ].
\end{equation}
I.e., from \eqref{eq:1a} and \eqref{eq:1b} we find that for any $\lambda \geq 1$ it holds
\begin{equation} \label{eq:3b}
\mathbb{E} [ \ee^{ - \lambda \omega}  ]  \leq \mathbb{E} [ \ee^{ (\lambda-1) \omega}  ] = \ee^{\alpha(\lambda - 1)}.
\end{equation}
Using the bounds \eqref{eq:3a} and \eqref{eq:3b} we get for the second term on the right hand side of \eqref{eq:2nL}:
\begin{equation} \label{eq:f2}
	\begin{aligned}
\sum\limits_{n=1}^\infty \int\limits_\veps^L (\omega \ast^k \omega)(x - 2nL)  \, \dd x &= \sum\limits_{n=1}^\infty \int\limits_{\veps-2nL}^{(1-2n)L} (\omega \ast^k \omega)(x)  \, \dd x \\
& \leq \sum\limits_{n=1}^\infty \int\limits_{-\infty}^{-(2n-1)L} (\omega \ast^k \omega)(x)  \, \dd x  \\
& \leq \sum\limits_{n=1}^\infty \ee^{k \alpha( nL)} \ee^{-2 (nL)^2}, 
	\end{aligned}
\end{equation} 
where on the last step we use the Chernoff bound for each term with $\lambda = nL + 1$.
Substituting \eqref{eq:f1} and \eqref{eq:f2} into \eqref{eq:2nL}, we see that
\begin{equation} \label{eq:C13}
	\int\limits_\veps^L \abs{C_1(x)}  \dd x \leq 2 \sum\limits_{n=1}^\infty \ee^{k \alpha( nL)} \ee^{-2 (nL)^2}.
\end{equation}

Moreover,
\begin{equation} \label{eq:C2}
\begin{aligned}
\int\limits_\veps^L \abs{C_2(x) }  \dd x =	&  \int\limits_\veps^L \, \int\limits_{-L}^L \omega(t_1) \ldots \int\limits_{-L}^L  \omega(t_{k-1})  r (x - \sum\nolimits_{i=1}^{k-1} t_i ) \, \dd t_1 \ldots \dd t_{k-1}  \, \dd x \\
	= &  \int\limits_{-L}^L \omega(t_1) \ldots \int\limits_{-L}^L  \omega(t_{k-1})   \int\limits_\veps^L r (x - \sum\nolimits_{i=1}^{k-1} t_i ) \, \dd x \, \dd t_1 \ldots \dd t_{k-1}.
\end{aligned}
\end{equation}
Clearly, for the inner factor in the integrand it holds by the Chernoff bound (setting $\lambda=L/2$)
$$
 \int\limits_\veps^L r (x - \sum\nolimits_{i=1}^{k-1} t_i ) \, \dd x \leq \int\limits_L^\infty \omega(t) \, \dd t \leq \ee^{\alpha(L/2)} \ee^{- \frac{L^2}{2}}.
$$
Thus, from \eqref{eq:C2} it follows that
\begin{equation} \label{eq:C22}
	\int\limits_\veps^L \abs{C_2(x) }  \dd x \leq \ee^{\alpha(L/2)} \ee^{- \frac{L^2}{2}}.
\end{equation}
Substituting \eqref{eq:C13} and \eqref{eq:C22} into \eqref{eq:C12}, we get
$$
\int\limits_\veps^L\abs{(\omega \circledast^k \omega - \widetilde{\omega} \circledast^k \widetilde{\omega})(x)} \, \dd x  \leq 
 \ee^{\alpha(L/2)} \ee^{- \frac{L^2}{2}} + 2 \sum\limits_{n=1}^\infty \ee^{k \alpha( nL)} \ee^{-2 (nL)^2}.
$$
\end{proof}
\end{lem}

\subsection{Error expansion with respect to $\Delta x $} \label{subsec:err_exp}

The purpose of this section is to show that the following assumption used in the main text holds (recall $\Delta x = 2L/n$): \\

There exists a constant $K$ independent of $n$ such that
\begin{equation} \label{eq:ass1}
	\begin{aligned}
\int\limits_\veps^L (1 - \ee^{\veps - s})(\widetilde{\omega} \circledast^k \widetilde{\omega} ) (s)  \, \dd s
		- \Delta x \sum\limits_{\ell=0}^{n-1}  \big(1 - \ee^{\veps - ( \ell \Delta x)} \big) C^k_\ell
= K \Delta x + O \big((\Delta x)^2 \big).
	\end{aligned}
\end{equation}

We motivate this assumption using the Euler--Maclaurin summation formula which gives the following expansion for the error of the Riemann sum formula (see~\cite[Ch.\;3.3]{stoer_book}).
\begin{lem}[Euler--Maclaurin formula]
Let $f \in C^{2m+2}[a,b]$. Let $N \in \mathbb{N}^+$ and denote $\Delta x =(b-a)/N$. Then,
\begin{equation*}
	\begin{aligned}
		\Delta x \sum\limits_{i=0}^{N-1} f(a + i \Delta x)  - \int_a^b f(x) \, \dd x = \Delta x \frac{f(a) - f(b)}{2} & + \sum\limits_{\ell=1}^m (\Delta x)^{2\ell} \frac{B_{2\ell}}{(2\ell)!} \big(f^{(2\ell-1)}(b) -  f^{(2\ell-1)}(a)   \big) \\
&		+ (\Delta x)^{2m+2} \frac{B_{2m+2}}{(2m+2)!} f^{(2m+2)}(\eta), \quad \eta \in [a,b],
	\end{aligned}
\end{equation*}
where $B_i$ is the $i$th Bernoulli number.
	
\end{lem}

Consider the discrete convolution vector $C^k$ as defined in Section 5.
By definition (summations periodic, indices modulo $n$),
\begin{equation*} 
		C_i^k = \Delta x \sum\limits_{j=0}^{n-1} \widetilde{\omega}(j \Delta x) C_{i-j}^{k-1}, \quad 
		C_i^2 = \Delta x \sum\limits_{j=0}^{n-1} \widetilde{\omega}(j \Delta x) \widetilde{\omega}(i \Delta x -j \Delta x),
\end{equation*}
If, instead, we consider the discrete convolutions
\begin{equation*} 
		\widehat{C}_i^k = \Delta x \sum\limits_{j=0}^{n-1} \omega (j \Delta x) C_{i-j}^{k-1}, \quad 
		\widehat{C}_i^2 = \Delta x \sum\limits_{j=0}^{n-1} \omega(j \Delta x) \omega(i \Delta x -j \Delta x),
\end{equation*}
then by the Euler--Maclaurin formula there clearly exist a constant $K$ independent of $n$ such that 
$$
\widehat{C}_i^k - (\omega \circledast^k \omega)(-L + i \Delta x ) = K \Delta x + O \big((\Delta x)^2 \big)
$$
for all $k=1,\ldots$ and $i=0,1,\ldots,n-1$.
Since the integrands in the convolution integrals of $\omega \circledast^k \omega$ are piecewise smooth
(we omit details here), it also has to hold
\begin{equation} \label{eq:Cik}
	C_i^k - (\widetilde{\omega} \circledast^k \widetilde{\omega})(-L + i \Delta x ) = K \Delta x + O \big((\Delta x)^2 \big)
\end{equation}
for some constant $K$ independent of $n$.

Using \eqref{eq:Cik} and the Euler--Maclaurin formula and the fact that the expressions in \eqref{eq:ass1} are piecewise smooth,
verifies the assumption \eqref{eq:ass1}.

\section{Auxiliary results} \label{sec:aux}

The following lemma is needed in the derivation of Newton's iteration.
\begin{lem} \label{lem:newton}
Let
$$
f(\veps) = \int_{\veps}^\infty (1- \ee^{\veps-s}) g(s) \, \dd s.
$$
Then,
$$
f'(\veps) = - \int_{\veps}^\infty \ee^{\veps-s} g(s) \, \dd s.
$$
\begin{proof}
Writing
$$
f(\veps) = \int_{\veps}^\infty g(s) \, \dd s - \ee^{\veps} \int_{\veps}^\infty \ee^{-s} g(s) \, \dd s
$$
and using the fundamental theorem of calculus and the chain rule, we see that
$$
f'(\veps) =  - g(\veps)  - \ee^{\veps} \int_{\veps}^\infty \ee^{-s} g(s) \, \dd s + \ee^{\veps} \cdot \ee^{-\veps} g(\veps) = - \int_{\veps}^\infty \ee^{\veps-s} g(s) \, \dd s.
$$
\end{proof}
\end{lem}

Recall that for the error analysis we consider the neighbouring relation $\sim_R$, i.e., we consider
the density function
$$
\omega(s) = \begin{cases}
f(g(s))g'(s), &\text{ if }  s > \log(1-q), \\
0, &\text{ otherwise},
\end{cases}
$$
where
\begin{equation} \label{eq:f_X}
f(t) = \frac{1}{\sqrt{2 \pi \sigma^2}} \, [ q \ee^{ \frac{-(t-1)^2}{2 \sigma^2}} + (1-q) \ee^{-\frac{t^2}{2 \sigma^2}} ],
\end{equation}
\begin{equation} \label{eq:Lminus}
g(s) = \sigma^2 \log \left( \frac{\ee^s - (1-q)}{q} \right) + \frac{1}{2}.
\end{equation}
The following lemmas which are be needed in the analysis of the approximation error.
\begin{lem} \label{lem:boundy}

For all $s \in ( \log (1-q), \infty)$:
$$
\omega(s) \leq \frac{\sigma}{q \sqrt{2 \pi}} \ee^{\frac{1}{\sigma^2}}.
$$

\begin{proof}
Consider first the case $s \in ( \log (1-q) , 0 ]$.
We see that then  $g(s) \in (-\infty,\frac{1}{2}]$ and
therefore
$$
\ee^{ - \frac{ (g(s) - 1)^2 }{ 2 \sigma^2 }  } \leq \ee^{ - \frac{ g(s)^2 }{ 2 \sigma^2 } }.
$$ 
Thus, 
\begin{equation} \label{eq:use1}
f(g(s)) \leq \frac{1}{\sqrt{2 \pi \sigma^2}} \ee^{ - \frac{ g(s)^2 }{ 2 \sigma^2 } }.	
\end{equation}
Moreover, for all $s \in ( \log (1-q) , 0 ]$,
\begin{equation} \label{eq:use2}
g'(s) = \frac{\sigma^2 \ee^s}{\ee^s - (1-q)} \leq \frac{\sigma^2}{\ee^s - (1-q)}.		
\end{equation}
Using \eqref{eq:use1} and \eqref{eq:use2}, we find that
\begin{equation} \label{eq:bound_x}
\omega(s) \leq \frac{\sigma}{\sqrt{2\pi}} \frac{\ee^{ - \frac{ g(s)^2 }{ 2 \sigma^2 } }	}{\ee^s - (1-q)}.
\end{equation}
We make the change of variables $x=g(s)$. Then,
$$
 \frac{1}{\ee^s - (1-q)} = q^{-1} \, \ee^{ \frac{-2x + 1}{2 \sigma^2} }
$$
and from \eqref{eq:bound_x} we see that
$$
\omega(s) \leq \frac{\sigma}{q \sqrt{2\pi}} \ee^{ \frac{- x^2 }{ 2 \sigma^2 } }	\ee^{ \frac{-2x + 1}{2 \sigma^2} } =
\frac{\sigma}{q \sqrt{2\pi}}   \ee^{ \frac{1}{ \sigma^2} }  \ee^{ \frac{-(x+1)^2 }{ 2 \sigma^2 }}
\leq \frac{\sigma}{q \sqrt{2\pi}}   \ee^{ \frac{1}{ \sigma^2} }
$$
which shows the claim for $s \in ( \log (1-q) , 0 ]$.

Assume next $s \geq 0$. Then,
$$
g'(s) = \frac{\sigma^2 \ee^s}{\ee^s - (1-q)} = \frac{\sigma^2}{1 - \frac{1-q}{\ee^s}} \leq \frac{\sigma^2}{q}.
$$
Since $f( g( s) ) \leq \frac{1}{\sqrt{2 \pi \sigma^2}}$, we see that when $s>0$,
\begin{equation*} 
\omega(s) \leq \frac{\sigma}{q \sqrt{2 \pi}}.
\end{equation*}
\end{proof}
\end{lem}

\begin{lem} \label{lem:boundy2}

For all $s \geq 1$ and $0<q\leq \tfrac{1}{2}$:
$$
\omega(s) \leq \sigma \ee^{ - \frac{ - ( \sigma^2 s + C)^2 }{2 \sigma^2} }, 
$$
where $C = \sigma^2 \log( \frac{1}{2q}) - \frac{1}{2}$.

\begin{proof}
Since $s\geq1$,
\begin{equation*} 
\ee^s - (1-q) \geq \frac{1}{2} \ee^s	
\end{equation*}
and therefore
$$
g(s) = \sigma^2 \log \left( \frac{\ee^s - (1-q)}{q} \right) + \frac{1}{2} \geq 
\sigma^2 s + C,
$$
where $C = \sigma^2 \log( \frac{1}{2q}) + \frac{1}{2}$. We see that $C \geq \frac{1}{2}$, since $0<q \leq \tfrac{1}{2}$.
Then also
$$
f(g(s)) \leq  \frac{1}{\sqrt{2 \pi \sigma^2}} \ee^{ - \frac{ - ( \sigma^2 s + C )^2 }{2 \sigma^2} },
$$
Furthermore, when $s>1$,
$$
g'(s) = \frac{\sigma^2 \ee^s}{\ee^s - (1-q)}  \leq 2 \sigma^2.
$$
Thus, when $s>1$,
$$
\omega(s) \leq \sigma \sqrt{ \frac{2}{\pi} } \ee^{ - \frac{ - ( \sigma^2 s + C_2 )^2 }{2 \sigma^2} }
\leq \sigma \ee^{ - \frac{ - ( \sigma^2 s + C_2 )^2 }{2 \sigma^2} }.
$$

\end{proof}
\end{lem}

\subsection{Bounds for derivatives}

\begin{lem} \label{lem:boundd}

Suppose $\sigma \geq 1$. For all $s \in ( \log (1-q), \infty)$:
$$
		\abs{\omega'(s)} \leq  4 \ee^{\frac{3}{\sigma^2}} \frac{\sigma^3}{q^2}
$$
and
$$
		\abs{\omega''(s)} \leq 11\ee^{\frac{9}{2 \sigma^2}} \frac{\sigma^3}{q^3}.
$$
\begin{proof}
Denote
$$
\omega(s) = f(g(s))g'(s),
$$
where
\begin{equation} \label{eq:g}
	g(s) =  \sigma^2 \log \left( \frac{\ee^s - (1-q)}{q} \right) + \frac{1}{2}
\end{equation}
and
$$
f(t) = \frac{1}{\sqrt{2 \pi \sigma^2}} \, [ q \ee^{ \frac{-(t-1)^2}{2 \sigma^2}} + (1-q) \ee^{-\frac{t^2}{2 \sigma^2}} ].
$$
Straightforward calculation shows that
\begin{equation} \label{eq:d1}
	\begin{aligned}
		\omega'(s) = f'(g(s)) (g'(s))^2 + f(g(s)) g''(s)
		\end{aligned}
\end{equation}
and
\begin{equation} \label{eq:d2}
	\begin{aligned}
		\omega^{(2)}(s) = f''(g(s)) (g'(s))^3 + 3 f'(g(s)) g''(s) g'(s) + f(g(s)) g^{(3)}(s).
		\end{aligned}
\end{equation}
Moreover,
\begin{equation} \label{eq:gd}
	\begin{aligned}
		g'(s) & = \frac{\sigma^2 \ee^s}{\ee^s - (1-q)}, \\
		g''(s) & = - \frac{\sigma^2 (1-q) \ee^s}{\big( \ee^s - (1-q) \big)^2}, \\
		g^{(3)}(s) &= \frac{\sigma^2 (1-q) \ee^s(\ee^s + (1-q))}{\big( \ee^s - (1-q) \big)^3}.
	\end{aligned}
\end{equation}
\textbf{Case $s \geq 0$.}
When $s \geq 0$, it holds
$$
 \frac{\ee^s}{\ee^s - (1-q)} = \frac{1}{1 - \frac{1-q}{\ee^s}} \leq \frac{1}{q}
$$
and from this inequality and expressions \eqref{eq:gd} it follows that
\begin{equation} \label{eq:ygeq0}
	\begin{aligned}
		\abs{g'(s)} & \leq \frac{\sigma^2}{q}, \\
		\abs{g''(s)} & \leq \frac{\sigma^2 }{q^2}, \\
		\abs{g^{(3)}(s)} & \leq \frac{2\sigma^2 }{q^3}. \\
	\end{aligned}
\end{equation}
Notice that when $s \geq 0$, $g(s) \geq \frac{1}{2}$.
By an elementary calculus, we find that when $\sigma \geq 1$, for $t \geq \frac{1}{2}$ it holds
\begin{equation} \label{eq:fds}
	\begin{aligned}
		f(t) &\leq \frac{1}{ \sigma  },\\
		f'(t) &\leq \frac{1}{\sqrt{2 \pi \sigma^2}} \frac{t}{\sigma} \ee^{- \frac{(t-1)^2}{2 \sigma^2}} \leq \frac{1}{\sigma} \\
		f''(t) &\leq \frac{1}{\sqrt{2 \pi \sigma^2}} \left(  \frac{t^2}{\sigma^2} + \frac{1}{\sigma^2}  \right) \ee^{- \frac{(t-1)^2}{2 \sigma^2}} \leq \frac{2}{\sigma^3}.
	\end{aligned}
\end{equation}
Substituting \eqref{eq:fds} and \eqref{eq:ygeq0} into \eqref{eq:d1} and \eqref{eq:d2} we find that
\begin{equation} \label{eq:geq_bound}
	\begin{aligned}
		\abs{\omega'(s)} &\leq 2 \frac{\sigma^3}{q^2}, \\
		\abs{\omega''(s)} & \leq 7 \frac{\sigma^3}{q^3}.
	\end{aligned}
\end{equation}
when $s \geq 0$. \\

\textbf{Case $s \in (\log(1-q,0))$}.
When $s \in (\log(1-q),0)$, from \eqref{eq:gd} it follows that
\begin{equation} \label{eq:gdb}
	\begin{aligned}
		\abs{g'(s)} & \leq \frac{\sigma^2}{\ee^s - (1-q)}, \\
		\abs{g''(s)} & \leq \frac{\sigma^2 }{\big( \ee^s - (1-q) \big)^2}, \\
		\abs{g^{(3)}(s)} & \leq \frac{2 \sigma^2  }{\big( \ee^s - (1-q) \big)^3}.
	\end{aligned}
\end{equation}
Consider next the five terms on the right hand sides of \eqref{eq:d1} and \eqref{eq:d2}. Consider first the term $f(g(s)) g''(s)$. By \eqref{eq:gdb}, we have the bound
\begin{equation} \label{eq:ft}
	f(g(s)) g''(s) \leq f(g(s)) \frac{\sigma^2}{\big(\ee^s - (1-q)\big)^2}.
\end{equation} 
Next, make the change of variables $x=g(s)$. Then, since (see \eqref{eq:g})
\begin{equation} \label{eq:cv}
    \frac{1}{\ee^s - (1-q)} = q^{-1} \, \ee^{ \frac{-2x + 1}{2 \sigma^2} },
\end{equation}
the bound \eqref{eq:ft} gives
\begin{equation} \label{eq:fb}
	f(g(s)) g''(s) \leq \sigma^2 q^{-2} f(x)  \ee^{ \frac{-4x + 2 }{2 \sigma^2} } \leq  
	\sigma^2 q^{-2}  \ee^{ \frac{3}{ \sigma^2} } \frac{ 1}{\sqrt{2 \pi \sigma^2}} \ee^{\frac{-(x+2)^2 }{2 \sigma^2} } \leq 
	\frac{ 1}{\sqrt{2 \pi }} \ee^{ \frac{3}{ \sigma^2} } \frac{\sigma}{q^2},
\end{equation}
as $f(t) \leq \frac{ 1}{\sqrt{2 \pi \sigma^2}} \ee^{\frac{-t^2 }{2 \sigma^2} }$ for $t \leq \frac{1}{2}$ and $g(s) \leq \frac{1}{2}$ for $s \in (\log(1-q,0))$.
With a similar technique, i.e., by using the change of variables $x=g(s)$ and \eqref{eq:cv}, we find after tedious calculation that
\begin{equation} \label{eq:sb}
	\begin{aligned}
		f'(g(s)) \big(g'(s)\big)^2 &\leq  3 \ee^{\frac{3}{\sigma^2}} \frac{\sigma^3}{q^2}, \\
		f''(g(s)) \big(g'(s)\big)^3 &\leq  6 \ee^{\frac{9}{2 \sigma^2}} \frac{\sigma^3}{q^3}, \\
		3 f'(g(s)) g''(s) g'(s) &\leq  4  \ee^{\frac{9}{2 \sigma^2}} \frac{\sigma^3}{q^3}, \\
		f(g(s)) g^{(3)}(s) &\leq \frac{ 1}{\sqrt{2 \pi}} \ee^{\frac{9}{2 \sigma^2}} \frac{\sigma}{q^3}.
	\end{aligned}
\end{equation}
Substituting \eqref{eq:fb} and \eqref{eq:sb} into \eqref{eq:d1} and \eqref{eq:d2} gives the bounds
\begin{equation} \label{eq:leq_bound}
	\begin{aligned}
		\abs{\omega'(s)} &\leq  4 \ee^{\frac{3}{\sigma^2}} \frac{\sigma^3}{q^2}, \\
		\abs{\omega''(s)} &\leq 11\ee^{\frac{9}{2 \sigma^2}} \frac{\sigma^3}{q^3}
	\end{aligned}
\end{equation}
for all $s \in (\log(1-q,0))$.

The claim follows from the bounds \eqref{eq:geq_bound} and \eqref{eq:leq_bound}.

\end{proof}
\end{lem}

\end{document}